\documentclass[conference]{IEEEtran}
\IEEEoverridecommandlockouts
\usepackage{cite}
\usepackage{amsmath,amssymb,amsfonts}
\usepackage{algorithmic}
\usepackage{graphicx}
\usepackage{textcomp}
\usepackage{xcolor}
\usepackage[ruled,linesnumbered,vlined]{algorithm2e}
\usepackage{balance}  
\usepackage{dsfont}
\usepackage{multirow}
\usepackage{xspace}
\usepackage{url}
\usepackage{CJK}
\usepackage{microtype}

\newcommand{\framework}{ExEA\xspace}

\newcommand{\revise}[1]{{\color{black}{#1}}}

\usepackage{xpatch}
\makeatletter
\def\changeBibColor#1{%
\in@{#1}{Anchor,llm_per,self-checker,lore,explainability,bert,kg_quality} 
\ifin@\color{black}\else\normalcolor\fi
}
\xpatchcmd\@bibitem
  {\item}
  {\changeBibColor{#1}\item}
  {}{\fail}
\xpatchcmd\@lbibitem
  {\item}
  {\changeBibColor{#2}\item}
  {}{\fail}
\makeatother

\def\BibTeX{{\rm B\kern-.05em{\sc i\kern-.025em b}\kern-.08em
    T\kern-.1667em\lower.7ex\hbox{E}\kern-.125emX}}

\begin{document}

\title{Generating Explanations to Understand and Repair Embedding-based Entity Alignment}

\author{\IEEEauthorblockN{Xiaobin Tian\IEEEauthorrefmark{2}, 
    Zequn Sun\IEEEauthorrefmark{2}\IEEEauthorrefmark{1}, 
    Wei Hu\IEEEauthorrefmark{2}\IEEEauthorrefmark{3}\IEEEauthorrefmark{1}\thanks{\IEEEauthorrefmark{1}Corresponding authors}}
    \IEEEauthorblockA{\IEEEauthorrefmark{2}{State Key Laboratory for Novel Software Technology, Nanjing University, Nanjing, China}\\
    \IEEEauthorrefmark{3}{National Institute of Healthcare Data Science, Nanjing University, Nanjing, China}\\
    xbtian.nju@gmail.com, sunzq@nju.edu.cn, whu@nju.edu.cn}
}

\maketitle
\thispagestyle{plain}
\pagestyle{plain}

\begin{abstract}
Entity alignment (EA) seeks identical entities in different knowledge graphs, which is a long-standing task in the database research.
Recent work leverages deep learning to embed entities in vector space and align them via nearest neighbor search.
Although embedding-based EA has gained marked success in recent years, 
it lacks explanations for alignment decisions.
In this paper, we present the first framework that can generate explanations for understanding and repairing embedding-based EA results.
Given an EA pair produced by an embedding model, we first compare its neighbor entities and relations to build a matching subgraph as a local explanation. 
We then construct an alignment dependency graph to understand the pair from an abstract perspective.
Finally, we repair the pair by resolving three types of alignment conflicts based on dependency graphs.
Experiments on a variety of EA datasets demonstrate the effectiveness, generalization, and robustness of our framework in explaining and repairing embedding-based EA results.
\end{abstract}


\section{Introduction}
Entity alignment (EA) is the task of finding identical entities in different knowledge graphs (KGs).
It can promote multi-source knowledge sharing and transfer to better support downstream applications.
EA is a long-standing task in the database and Semantic Web fields.
Early work on EA mainly uses string matching to compute entity similarity and infers EA via similarity propagation~\cite{sf,PARIS}.
The development of deep learning techniques motivates recent work to represent entities as vector representations (i.e., embeddings) and find EA using the nearest neighbor search in vector space \cite{MTransE}.
Typically, embedding-based EA models use shallow or deep neural networks, e.g., TransE~\cite{TransE} or graph convolutional networks (GCNs) \cite{GCN}, to encode entities and use the output vector representations to compute the similarity between entities.
In the alignment inference phase, given a source entity $e_1$ to be aligned, the model usually greedily selects the most similar entity $e_2$ in the target KG as the counterpart.

Recently, embedding-based EA has attracted increasing attention~\cite{tkde_ea,vldbj_ea}. 
Hundreds of studies in the past several years have explored advanced representation learning techniques,
such as relational GCNs~\cite{RelationalEA}, multi-hop GCNs~\cite{AliNet}, and dual GCNs~\cite{DualAMN}, to train informative entity embeddings for similarity computation. 
Others have improved EA inference via bi-directional $k$NN search~\cite{MRAEA}, reinforcement learning~\cite{icde_ea_rl,tois_ea_rl}, and holistic matching~\cite{BootEA,OpenEA}.
However, embedding-based EA models are unexplainable, like other deep learning models.
We cannot discern the reasoning behind their output.
Yet, no effort has been made to provide explanations for EA.
Understanding embedding-based EA results can offer profound insight into the distinctions between EA models.
Additionally, EA explanations can act as background knowledge to assist users in judging the reliability of EA results.
Therefore, we study EA explanations in this paper.

\smallskip\noindent\textbf{Challenges.} Explaining embedding-based EA results is a challenging task.
The absence of prior research presents the first challenge.
Existing machine learning algorithm explanation approaches strive to discover the most influential input features that can change the result \cite{LIME}.
In the case of EA, an alignment output (e.g., $e_1 \equiv e_2$) is determined by the features (i.e., relation triples) of both entities.
Hence, existing explanation approaches cannot be directly applied to embedding-based EA.
The second challenge lies in the complexity of embedding-based EA models.
GCN-based EA models \cite{DualAMN} usually consider high-order triples for alignment inference. It is time-consuming to enumerate all possible combinations of multi-order triples and thus difficult to measure the importance of each triple in determining the EA result. 
The third challenge is how to use explanations to repair model results.
To the best of our knowledge, this is the first attempt that takes explanation feedback into account, making it possible for the explanation generation methods to be used in real-world situations.

\smallskip\noindent\textbf{Solution.} To resolve the above challenges, we present the \framework framework in this paper. 
Based on the heuristic cognition of the EA task, \framework regards the generation of EA explanations as the generation of semantic matching subgraphs, which can take into account the characteristics of two entities simultaneously by matching semantically consistent triples around them.
In order to provide comprehensive explanations, \framework treats all the matching triples as important features.
This allows us to bypass the process of evaluating the influence of combined features on the model. 
Not only does this save a considerable amount of time, but it also ensures that the generated explanations remain unbiased and faithful to the original model, unaffected by any biases that might arise during the estimation of feature importance.
After obtaining the explanation, we devise an alignment dependency graph structure (ADG) to gain deeper insights into the explanation. 
Using this graph, we can calculate the explanation confidence, which plays a crucial role in resolving conflicts in EA results and repairing them effectively.

\smallskip\noindent\textbf{Contributions.} Our main contributions are listed as follows:

\begin{itemize}
    \item We provide a definition of the explanation for embedding-based EA, which considers the fidelity of EA results.
    
    \smallskip\item We propose \framework, an extensible framework that is able to generate \revise{high quality} explanations for a given EA model. It can also repair the EA results. \framework can be applied to any embedding-based EA model.
    
    \smallskip\item We incorporate four representative EA models into \framework and evaluate their effectiveness on benchmark datasets.

    \smallskip\item We make several findings on embedding-based EA. First, simple models can also achieve high accuracy by effectively repairing alignment conflicts. Second, the one-to-many alignment is the most common and most influential conflict. Third, there is some alignment that cannot be identified based on structures in the benchmark \cite{JAPE}. 
\end{itemize}

\smallskip\noindent\textbf{Outline.} 
\revise{In Section~\ref{sect:prelim}, we introduce notations, definitions, and the pipeline of \framework.
We describe our explanation generation method in Section~\ref{sect:explanation} and our EA repair method in Section~\ref{sect:repair}.}
We conduct experiments in Section~\ref{sect:exp} and discuss related work in Section\ref{sect:related_work}.
Finally, we conclude this paper in Section~\ref{sect:concl}.

\section{Preliminaries}\label{sect:prelim}

\revise{\subsection{Notation}}
\revise{In this paper, we use uppercase letters with calligraphic font to denote the element set of KGs, such as the entity set $\mathcal{E}$ and the EA set $\mathcal{A}$.
We use italic lowercase letters to denote entity or relation variables, such as entity $e$.
We use bold letters to represent vectors or matrices, such as the entity embedding $\mathbf{e}$ or similarity matrix $\mathbf{M}$.
Function names in our framework are represented in typewriter font, such as function $\texttt{weight}()$.}

\revise{\subsection{Definitions}}
\revise{\smallskip\noindent\textbf{Knowledge graph.}
A KG is defined as $\mathcal{K}=(\mathcal{E}, \mathcal{R}, \mathcal{T})$, where $\mathcal{E}$ and $\mathcal{R}$ denote the entity set and relation set, respectively.
$\mathcal{T} \subseteq \mathcal{E} \times \mathcal{R} \times \mathcal{E}$ denotes the set of relation triples. 
Specifically, a relation triple is in the form of (subject, \textit{relation}, object), where the subject and object are entities. 
}

\smallskip\noindent\textbf{Entity alignment.}
Given a source KG $\mathcal{K}_1$ and a target KG $\mathcal{K}_2$,
EA is the task of finding the equivalence relations between entities $\mathcal{E}_1$ and $\mathcal{E}_2$, 
i.e., $\mathcal{A}=\{(e_1, e_2)\in \mathcal{E}_1 \times \mathcal{E}_2\,|\, e_1 \equiv e_2 \}$, where $\equiv$ denotes the ``\text{owl:sameAs}'' relation.
In most cases, a small set of seed EA \revise{$\mathcal{A}_{train}$} is provided as training data, and the task is to find the remaining alignment \revise{$\mathcal{A}_{res}=\{(e'_1, e'_2)\in \mathcal{E}'_1 \times \mathcal{E}'_2\,|\, e'_1 \equiv e'_2 \}$, where $\mathcal{E}'_1$ and $\mathcal{E}'_2$ denote the entity sets to be aligned of the two KGs.}

\smallskip\noindent\textbf{Explanation for entity alignment.}
We focus on providing post-hoc explanations for specific results of EA models. 
\revise{Given the EA pair ($e_1 \equiv e_2$) produced by a model, 
we define the triples within \revise{$h$} hops from $e_1$ and $e_2$ by $\mathcal{T}_{e_1}$ and $\mathcal{T}_{e_2}$, respectively.
Larger \revise{$h$} may lead to more irrelevant triples, so we typically choose $\revise{h} \leq 2$.
We define $\mathcal{T}_{(e_1, e_2)}=\mathcal{T}_{e_1} \cup \mathcal{T}_{e_2}$ as the candidate triples for explanation generation.
Our objective is to identify the smallest subset $\mathcal{T}^*_{(e_1, e_2)}\subset\mathcal{T}_{(e_1, e_2)},\mathcal{T}^*_{(e_1, e_2)} \neq \emptyset$, such that if we remove $\mathcal{T}_{(e_1, e_2)}-\mathcal{T}^*_{(e_1, e_2)}$ from KGs, the EA model can still make the corresponding prediction ($e_1 \equiv e_2$)\revise{\cite{explainability}}.
The subgraph formed by the entities and relations in the selected triples serves as the explanation for ($e_1 \equiv e_2$).
We use the metric fidelity~\cite{explainability} to measure the quality of EA explanations.
Higher fidelity indicates a higher quality of explanations.
}

\revise{\smallskip\noindent\textbf{Horn rule.}
A Horn rule has a head atom and one or more body atoms connected by logical conjunction, typically represented as $B \rightarrow (x, r, y)$, with $B$ representing the body.
An atom is a triple that involves variables in at least one of the subject or object positions. 
For instance, in the atom $(x, r, y)$, $x$ and $y$ are variables linked by relation $r$.
}
\begin{figure}[!t]
	\centering
	\includegraphics[width=\linewidth]{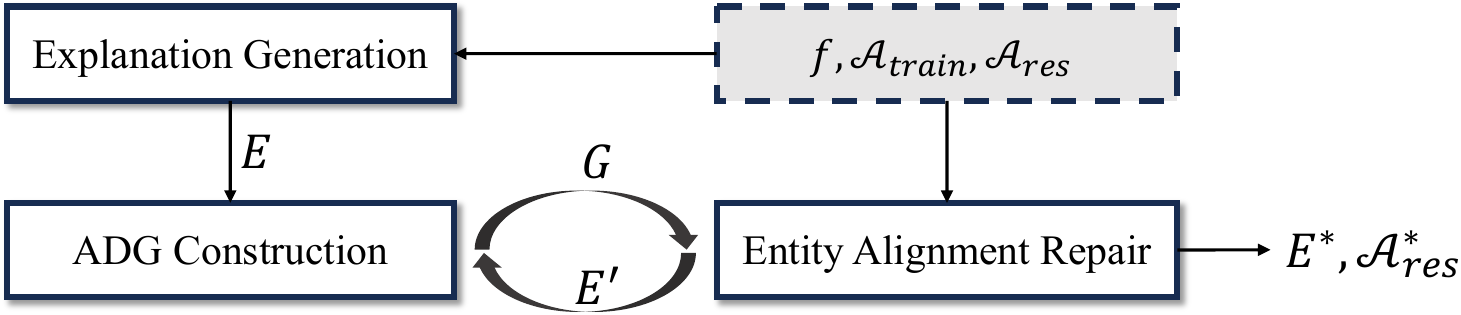}
	\caption{\revise{Overview of the \framework framework.}}
	\label{fig:framework}
\end{figure}

\revise{\subsection{Framework Overview}}
\label{sect:overview}
Fig.~\ref{fig:framework} shows the pipeline of \framework, which has three modules: \emph{explanation generation}, \emph{ADG construction}, and \emph{EA repair}.

\revise{\smallskip\noindent\textbf{Input.}
The input of \framework consists of a trained EA model $f$ and its predicted EA results $\mathcal{A}_{res}$ for two KGs $\mathcal{K}_1$ and $\mathcal{K}_2$.}

\begin{figure*}[!t]
\centering
\includegraphics[width=\textwidth]{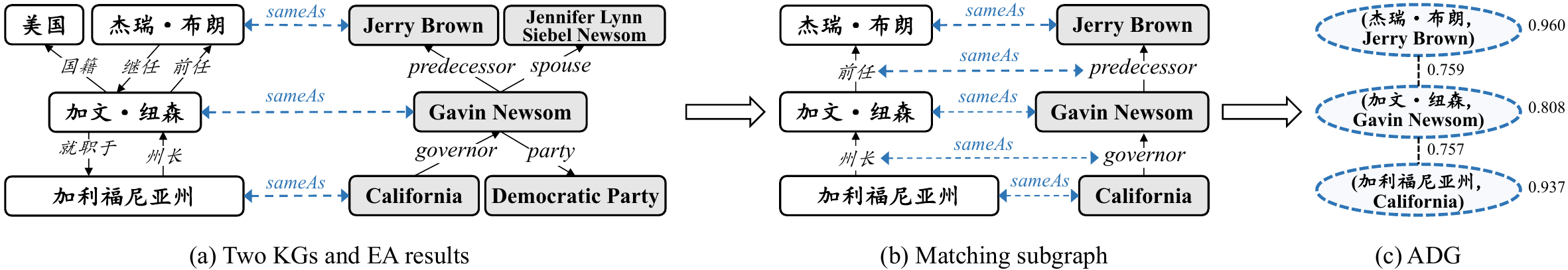}
\caption{\revise{Illustration of explanation generation and ADG construction. In ADG, each edge has a weight indicating how a node can influence others. The edge weight is calculated based on the functionality of relations. For example, $0.759 = \min\big(\texttt{ifunc}(\text{\begin{CJK}{UTF8}{gbsn}前任\end{CJK}}), \texttt{ifunc}(predecessor)\big)$. The node in ADG denotes an EA pair, which has a match confidence calculated based on neighbor alignment. For example, $0.808 = \texttt{sigmod}(0.960\times0.759 + 0.937\times0.757)$}.}
\label{fig:EGRE}
\end{figure*}

\revise{\smallskip\noindent\textbf{Pipeline.}} 
First, \framework generates a semantic matching subgraph for the EA pair as a local explanation denoted by $E$ (Section~\ref{sect:explain_gene}). 
Then, \framework uses matched entities or relations in local explanations to construct an alignment dependency graph (ADG) that explains the output from an abstract perspective (Section~\ref{sect:adg}).
Finally, \framework uses ADGs to detect three types of EA conflicts and repair them (Section~\ref{sect:repair}).

\revise{\smallskip\noindent\textbf{Output.}}
\framework outputs the finally repaired EA results $\mathcal{A}^*_{res}$ with corresponding explanations $E^*$.

\medskip\section{\revise{EA Explanation and Dependency Construction}}\label{sect:explanation}

In this section, we introduce EA explanation generation and alignment dependency graph construction.

\subsection{\revise{Explanation Generation}}\label{sect:explain_gene}
\label{eg}
Given the EA pair ($e_1 \equiv e_2$) predicted by an EA model, our explanation generation method seeks to efficiently identify the relation triples that support the model to make the prediction. 

\smallskip\noindent\textbf{Motivation.}
Following existing work on explaining machine learning models,
we can design a naive method for explaining embedding-based EA, which typically uses each individual relation triple as the search unit and investigates the impact of different triple combinations on EA predictions. 
However, this method suffers from several flaws.
First, it is challenging to find reliable explanations.
The EA task requires considering the relation triples of two entities simultaneously. 
If analyzing them separately, it is difficult to capture associations between these triples, thereby undermining \revise{explanation quality~\cite{LIME}}.
Second, this method has exponential time complexity since it needs to analyze all combinations of relation triples\revise{\cite{Shapley}}. 
To address these challenges, we leverage the heuristic understanding of EA: 
\textit{two entities are aligned because their relation triples share similar semantics}.
Motivated by this, we model EA explanations as semantically matching subgraphs, as shown in Fig.~\ref{fig:EGRE} (a) and (b), thereby avoiding unnecessary search.

\smallskip\noindent\textbf{Semantic matching subgraph generation.} Specifically, it has two steps: \textit{match neighbor entities} and \textit{match relation paths}, as shown in Fig.~\ref{fig:EGRE} (b).
In the first step, considering that the explanation needs to be faithful to the EA results, we match the neighbors of $e_1$ and $e_2$ that are predicted to be aligned by the model or are themselves in seed alignment.
In the second step, we collect relation paths between the matched neighbor entities and the central entities and calculate path representations based on the involved relation and entity embeddings.
\revise{If the EA model captures relations like MTransE~\cite{MTransE}, we directly use their relation embeddings.
If the EA model does not learn relation embeddings, as is the case with GCN-Align~\cite{GCNAlign}, we adopt a method inspired by TransE~\cite{TransE} that computes relation embeddings through entity embedding translation:}
\begin{align}
\mathbf{r}=\frac{1}{|\mathcal{T}_r|}\sum_{(s,r,o)\in \mathcal{T}_r}(\mathbf{e}_s - \mathbf{e}_o),
\end{align}
where $\mathcal{T}_r$ is the set of triples containing relation $r$.
$\mathbf{e}_s$ and $\mathbf{e}_o$ denote the embeddings of entities $s$ and $o$, respectively.

Then, given a relation path from entity $e_1$ to its neighbor $e_{n}'$, i.e., $p = (e_1, r_1, e_{1}', r_2, e_{2}',\dots, r_{n}, e_{n}')$,
it is represented by:
\begin{align}
\mathbf{p}=\frac{\mathbf{e}_1 + \sum_{i=1}^{n-1}\mathbf{e}_i'}{n} \oplus \frac{\sum_{i=1}^{n}\mathbf{r}_i}{n}.
\end{align}

Finally, we perform bidirectional matching over path embeddings to match subgraphs.
An entity may have multiple relation paths with its neighbor entities.
\revise{Let $P_1 = \{p_1^i\}_{i=1,2,\dots,n_1}$ and $P_2 = \{p_2^j\}_{j=1,2,\dots,n_2}$ denote the path sets of $e_1$ and $e_2$, respectively.
For each ${p}_1\in P_1$, we identify ${p}_2^j\in P_2$  that exhibits the highest embedding similarity to it, i.e.,
$j=\arg \max_{j\in {1,2,\dots,n_2}}\cos(\mathbf{p}_1, \mathbf{p}_2^j)$.
In a similar way, we can find the most similar path ${p}_i\in P_1$ for each ${p}_2\in P_2$.
If there are two paths ${p}_1\in P_1$ and ${p}_2\in P_2$ that are the most similar to each other, we match the entities and relations along the two paths, resulting in semantically matched triples.
These triples form a semantic matching subgraph, as shown in Fig.~\ref{fig:EGRE} (b), which represents the explanation for the EA pair.}

\subsection{\revise{Alignment Dependency Graph Construction}}
\label{sect:adg}
Given the obtained explanation for each EA result, the next step is to delve deeper into understanding how each element of the explanation influences the EA effectiveness. 
We introduce an alignment dependency graph (ADG) to depict the influence.

\revise{\smallskip\noindent\textbf{Node construction.}} We merge two matched entities into a node and record their embedding similarity as the influence of the node in the explanation, as shown in Fig.~\ref{fig:EGRE} (c).
The node formed by the two entities to be explained is referred to as the central node, \revise{such as the node (\begin{CJK}{UTF8}{gbsn}加文$\cdot$纽森\end{CJK}, Gavin Newsom) in Fig.~\ref{fig:EGRE} (c)}. Other nodes are referred to as the neighbor nodes.

\smallskip\noindent\textbf{\revise{Edge construction.}} 
Then, we establish edges between the central node and its neighbor nodes through the matched relation paths.
An edge indicates that there are two matched relation paths between entities of the central and neighbor nodes.
Each edge has a weight that quantifies the impact of the neighbor node on the central node.
\revise{Inspired by~\cite{AliNet}}, we categorize edges based on the lengths of \revise{their} relation paths. 
\begin{itemize}
    \item If both relation paths in the edge have a length of one, it means that the entities in the neighbor nodes influence the entities in the central node through direct relation. 
    We refer to the edge as \emph{strongly influential}. 
    \item If only one of the two relation paths in the edge has a length of one, it indicates that there is only one entity in the neighbor node that directly affects one entity in the central node. 
    We label the edge as \emph{moderately influential}. 
    \item If the lengths of both relation paths in the edge are greater than one, it implies that the entities in the neighbor nodes have no direct relation that can affect the entities in the central node. 
    We term the edge as \emph{weakly influential}.
\end{itemize}
To differentiate the influence on the central node across various edge types, we employ distinct weight calculation methods. 

\smallskip\noindent\textbf{\revise{Edge weight computation.}}
We follow PARIS~\cite{PARIS} and calculate the weight of \emph{strongly influential} edges based on the relation functionality or inverse functionality.
If the central entity $e_1$ is the head entity of the path $p_1$, e.g., $p_1=(e_1$, $r_1$, $e_1')$,
the path weight is calculated using inverse functionality: 
\begin{align}
{\texttt{weight}}(p_1) = {\texttt{ifunc}}(r_1).
\end{align}
Otherwise, we use relation functionality. 
For example, if the other central entity $e_2$ is the tail entity of the path $p_2$, i.e., $p_2=(e_2', r_2, e_2)$, this path weight is calculated by
\begin{align}
{\texttt{weight}}(p_2) = {\texttt{func}}(r_2).
\end{align}

Then, for a \emph{strongly-influential} edge $l_s$ containing matched relation paths $p_1$ and $p_2$, its final weight is calculated as
\begin{align}
{\texttt{weight}}(l_s) = \texttt{min}\big({\texttt{weight}}(p_1), {\texttt{weight}}(p_2)\big).
\end{align}
\revise{This decision is driven by the consideration that the results of EA models inevitably have errors, and choosing the relatively low weight can reduce the potential impact of such inaccuracies.}

For \emph{moderately-influential} edges, the weight calculation is analogous. 
The weight of directly affected relation paths is determined based on relation functions. 
For a long path $p_l$ with indirect influence, we posit that its weight is collectively determined by its direct paths $p_1'=(e_1, r_1, e_1')$, $p_2' = (e_1', r_2, e_2')$, $\dots$, $p_n' = (e_{n-1}', r_{n}, e_n')$.
Its weight is calculated by:
\begin{align}
\resizebox{.9\columnwidth}{!}{$
{\texttt{weight}}(p_l)={\texttt{weight}}(p_1') \times {\texttt{weight}}(p_2') \times\cdots\times {\texttt{weight}}(p_n').$}
\end{align}

Then, for a \emph{moderately-influential} edge $l_m$ containing relation path $p_d$ with direct influence and relation path $p_l$ with indirect influence, its final weight is calculated as
\begin{align}
{\texttt{weight}}(l_m) = \alpha \times \min\big({\texttt{weight}}(p_d), {\texttt{weight}}(p_l)\big),
\end{align}
where $\alpha\leq 1$ is a hyper-parameter, which allows us to further differentiate the significance of various edges.

For \emph{weakly-influential} edges, their relation paths exhibit indirect influence, indicating that the neighbor nodes only have a minimal impact on the central node. 
Therefore, we directly assign a small predefined weight to represent their influence.

\revise{\smallskip\noindent\textbf{Node confidence computation.}}
We define node confidence as the likelihood that the EA pair under the explanation subgraph is valid. 
\revise{The confidence of the central node is the weighted aggregation of its neighbor node influence}:
\begin{align}
\resizebox{.9\columnwidth}{!}{$
c=\texttt{sigmoid} \Big(\sum\limits_{i=1}^{|\mathcal{N}|}\big(\sum\limits_{j=1}^{|\mathcal{L}_i|}{\texttt{weight}}(l_{ij})\big)\texttt{I}(n_i)\Big),
l_{ij}\in\mathcal{L}_i,n_i \in \mathcal{N},$}
\end{align}
where $\mathcal{N}$ denotes the set of neighbor nodes.
$\mathcal{L}_i$ denotes the set of edges between the $i$-th neighbor node $n_i$ and the central node.
$\texttt{I}(n_i)$ represents the influence of the two entities in the neighbor node $n_i$.
In practice, \emph{strongly-influential} edges are sufficient to represent the confidence of an explanation in most cases. 
Calculating other types of edges may increase time complexity and introduce noise. 
Therefore, we employ adaptive weights during aggregation to adjust the treatment of different edge types.
The modified confidence is calculated by:
\begin{align}
\resizebox{.9\columnwidth}{!}{$
c =\texttt{sigmoid} \Big(c_s + \mathds{1}(c_s < \theta) \times \big(c_m + \mathds{1}(c_m < \gamma) \times c_w\big)\Big),$}  \label{conf_compute}
\end{align}
where $c_s$, $c_m$, and $c_w$ denote the aggregation results of \emph{strongly-influential}, \emph{moderately-influential}, and \emph{weakly-influential} edges, respectively.
$\mathds{1}(\cdot)$ is an indicator function.
$\theta$ and $\gamma$ are hyper-parameters.
Given this, we can dynamically determine whether to proceed with further calculations based on current results, thereby saving time and minimizing potential noise.
\begin{figure*}[!t]
\centering
\includegraphics[width=\textwidth]{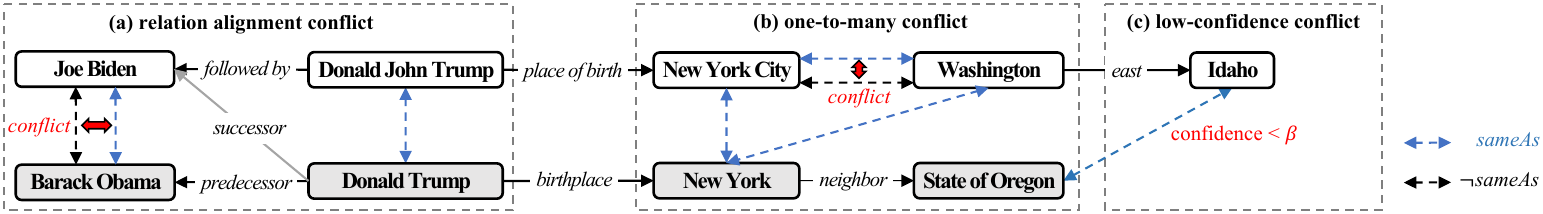}
\caption{\revise{Illustration of EA conflicts. The blue dotted arrow denotes the predicted ``sameAs'' relation of an EA model while the black dotted arrow denotes the inferred ``$\neg$ sameAs'' relation based on our method. $\beta$ is a predefined confidence threshold.}}
\label{fig:conflict1}
\end{figure*}

\section{\revise{Entity Alignment Repair}}\label{sect:repair}
We repair EA results by resolving three types of conflicts.

\subsection{\revise{Relation Alignment Conflicts}}
\revise{These conflicts indicate that the EA pair in the central node of an ADG has $\lnot sameAs$ relation inferred by their matched triples, like the pair (Joe Biden, Barack Obama) in Fig.~\ref{fig:conflict1} (a).}

\smallskip\noindent\textbf{\revise{$\mathbf{\lnot}$\textbf{\textit{sameAs}} relation mining.}} \revise{To detect these conflicts, we need to obtain the relation alignment between two KGs and mine the $\neg sameAs$ relation between entities.
To mine relation alignment, if relation names are available, we employ a pre-trained language model, such as BERT~\cite{bert}, to encode them and obtain relation embeddings. 
Otherwise, we can get relation embeddings from the original EA model.
Then, we use greedy matching to find aligned relations with the highest embedding similarity to each other, such as the relation alignment (\textit{followed by}, \textit{successor}) in Fig.~\ref{fig:conflict1} (a).}
Regarding the inference rules for the $\neg sameAs$ relation, 
\revise{
we are motivated by the idea that if an entity has two different relations and the corresponding tail entities are usually different,
we can generate the corresponding $\neg sameAs$ relation inference rule.
Specifically, given two different relations $r_1$ and $r_1'$ (i.e., not aligned), if triples $(e_1, r_1, e_1')$ and $(e_1, r_1', e_1')$ do not exist simultaneously for any entities $e_1$ and $e_1'$,
we can establish a candidate rule: $(x, r_1, y) \land (x, r_1', z) \land (r_1 \neg sameAs, r_1') \rightarrow (y, \neg sameAs, z)$.
However, this method may result in a large number of rules, most of which may be wrong or useless.}
To reduce the number of obtained rules,
we introduce an additional condition: 
\revise{we require the existence of real rule instances, e.g., at least one instance like $(e_1, r_1, e_1') \land (e_1, r_1', e_1'') \land (r_1, \neg sameAs, r_1') \rightarrow (e_1, \neg sameAs, e_1'')$ holds for relations $r_1$ and $r_1'$ in the KG.
}

\smallskip\noindent\textbf{\revise{Cross-KG triple construction.}} \revise{Given EA results, we generate cross-KG triples \cite{BootEA} by swapping aligned entities or relations in their triples.
Cross-KG triples allow us to reason over different KGs to find EA conflicts.
Given an EA pair $(e_1, e_2)$ and their triples $(e_1, r_1, e_1')$ and $(e_2, r_2, e_2')$,
we can generate the following cross-KG triples: $(e_2, r_1, e_1')$ and $(e_1, r_2, e_2')$.
To reduce the number of cross-KG triples for effective and efficient reasoning, 
we only generate cross-KG triples for the entities that have \emph{strongly-influential} edges in ADGs.
For instance, in Fig.~\ref{fig:conflict1} (a), (Donald Trump, \textit{successor}, Joe Biden) is a cross-KG triple, where entity Joe Biden is from $\mathcal{K}_1$ while entity Donald Trump and relation \textit{successor} are from $\mathcal{K}_2$.}

\smallskip\noindent\textbf{\revise{Conflict detection and repair.}} \revise{Relation alignment conflicts arise when the mined ``$\neg sameAs$'' rules can infer from cross-KG triples that the two entities in the central node of an ADG are not the same. 
Take Fig.~\ref{fig:conflict1} (a) as an example. 
Based on the mined rules, relation alignment, and cross-KG triples, we can infer that
(Donald John Trump, \textit{followed by}, Joe Biden) $\land$ (Donald John Trump, \textit{sameAs}, Donald Trump) $\land$ (\textit{followed by}, \textit{sameAs}, \textit{successor}) $\rightarrow$ (Donald Trump, \textit{successor}, Joe Biden), and then get that
(Donald Trump, \textit{successor}, Joe Biden) $\land$ (Donald Trump, \textit{predecessor}, Barack Obama) $\land$ (\textit{successor}, $\neg sameAs$, \textit{predecessor}) $\rightarrow$ (Joe Biden, $\neg sameAs$, Barack Obama).
However, the ADG tells us that (Joe Biden, $sameAs$, Barack Obama), where a conflict occurs.}
Note that both our relation alignment and the mined $\neg sameAs$ rules may contain errors. 
To address this issue, we define the relation alignment conflict as a soft conflict and weaken the influence of potentially erroneous EA pairs in ADGs.
We delete the neighbor nodes whose entities are inferred to be misaligned to obtain more reliable explanations and recalculate the explanation confidence, assisting in the subsequent EA repair progress.

\smallskip\noindent\textbf{\revise{Analysis.}}
\revise{In each ADG construction, we restrict the number of relation triples $\mathcal{T}_{\text{n}}$ surrounding entities when detecting relation alignment conflicts.
In each detection, the time consumption is affected by two aspects:
one is judging if a relation pair is in the relation alignment set $\mathcal{A}_{r}$ and the other entails evaluating if cross-KG triples can satisfy a rule in the $\neg sameAs$ rules set $\mathcal{S}$.
So, in each ADG construction, the time complexity of repair is $O\big(|\mathcal{T}_{\text{n}}|\times (\log(|\mathcal{A}_{r}|) + \log(|\mathcal{S}|)) \big)$. 
In practice, both $|\mathcal{A}_{r}|$ and $|\mathcal{S}|$ are small and $|\mathcal{T}_{\text{n}}|$ is restricted at a constant level.
}

\begin{algorithm}[!tb]
\caption{EA repair for one-to-many conflicts}
\label{alg:one_to_many}
{\small
\KwIn{Seed EA \revise{$\mathcal{A}_{train}$}; EA results \revise{$\mathcal{A}_{res}$}; 
pairwise similarity matrix \revise{$\mathbf{M}$} between unaligned source and target entities in descending order; 
number of candidate entities \revise{$k$}.}
\KwOut{Repaired EA $\mathcal{A}^*$.}
$\revise{\mathcal{E}'_{1}},\revise{\mathcal{A}^*}  \leftarrow$ \texttt{OnetoOne}($\mathcal{A}_{train}$, $\mathcal{A}_{res}$)\;
\While {$\texttt{\tt len}(\revise{\mathcal{E}_1'}) > 0$}{
    $lastLen \leftarrow \texttt{\tt len}(\revise{\mathcal{E}_1'}); \revise{\mathcal{E}'_{1new}} \leftarrow \texttt{\tt set}()$\;
    \For{$e_1 \in \revise{\mathcal{E}_1'}$}{
        Get the index $i$ of $e_1$\;
        \For{$j$ $\leftarrow 0$ \KwTo $k - 1$}{
            Get the entity $e_2$ indexed by $\revise{\mathbf{M}}[i][j]$\;
            \If{$e_2$ is not aligned}
                {
                $\revise{\mathcal{A}^*}.\texttt{\tt add}((e_1, e_2))$; \textbf{break}\;
                }
            \Else{
                Get the entity $e_1'$ aligned with $e_2$\;
                $x_1\leftarrow\texttt{Exp}(e_1, e_2, \revise{\mathcal{A}^*},\revise{\mathcal{A}_{train}})$\;
                $x_2 \leftarrow \texttt{Exp}(e_1', e_2, \revise{\mathcal{A}^*},\revise{\mathcal{A}_{train})}$\; 
                $g_1 \leftarrow \texttt{ADGConstruct}(x_1)$\;
                $g_2 \leftarrow \texttt{ADGConstruct}(x_2)$\;
                \If{$g_1.conf > g_2.conf$}
                {
                    $\revise{\mathcal{A}^*}.\texttt{\tt add}((e_1, e_2)); \revise{\mathcal{A}^*}.\texttt{\tt remove}((e_1', e_2))$;\\
                    $\revise{\mathcal{E}'_{1new}}.\texttt{\tt add}(e_1')$; \textbf{break}\;
                }
            }
        }
        \lIf{$e_1$ is not aligned}{$\revise{\mathcal{E}'_{1new}}.\texttt{\tt add}(e_1)$}  
    }
    $\revise{\mathcal{E}_1'} \leftarrow \revise{\mathcal{E}'_{1new}}$; \\
    \lIf{$\mathrm{len}(\revise{\mathcal{E}_1'}) \geq lastLen$}{\textbf{break}}
}}
\end{algorithm}

\subsection{\revise{One-to-Many Conflicts}}
These conflicts mainly arise when the EA results contradict the assumption that all entities in a KG are distinct.

\smallskip\noindent\textbf{\revise{Conflict detection.}}
\revise{In the predicated EA pairs, if there are multiple entities in the source KG $\mathcal{K}_1$ aligned with the same entity in the target KG $\mathcal{K}_2$, they will cause a conflict.}
For instance, assuming that $e_1$ and \revise{$e_1'$} are entities in \revise{$\mathcal{K}_1$}, since entities in a KG are different, we have \revise{($e_1, \neg sameAs, e_1'$)}. 
If both $e_1$ and $e_1'$ are predicted to be aligned with $e_2$ in \revise{$\mathcal{K}_2$ by an EA model, 
we have ($e_1, sameAs, e_2$) and ($e_1', sameAs, e_2$). }
Due to the transitivity of $sameAs$, we can deduce \revise{($e_1, sameAs, e_1'$)}, which contradicts the initial condition that \revise{($e_1, \neg sameAs, e_1'$)}.
\revise{As illustrated in Fig.~\ref{fig:conflict1} (b), New York City and Washington have the conflict caused by one-to-many EA pairs.}

\smallskip\noindent\textbf{\revise{Conflict repair.}}\revise{
Algorithm \ref{alg:one_to_many} outlines the repair process.
We first resolve one-to-many conflicts (Line 1).
We select the EA pair with the highest explanation confidence as the correct one and remove others.
As a result, we get the set of unaligned source entities  $\mathcal{E}'_{1}$ and the one-to-one EA set $\mathcal{A}^*$.
We iteratively realign the entities in $\mathcal{E}'_{1}$ (Line 2--21).
At the beginning of each iteration, 
we initialize a set $\mathcal{E}'_{1new}$ to collect the entities that are still not aligned in this iteration (Line 3).
Next, we retrieve the top-$k$ target entities similar to each unaligned source entity $e_1$ (Lines 4--7).
If the selected target entity $e_2$ has not been aligned previously, it will be aligned to $e_1$ (Lines 8--9).
Otherwise, if $e_2$ has an aligned entity $e_1'$, we compare the explanation confidences of $(e_1, e_2)$ and $(e_1', e_2)$, and choose the one with higher confidence as the correct alignment (Lines 11--18).
If none of the top-$k$ target entities can be aligned to $e_1$, we add $e_1$ to $\mathcal{E}'_{1new}$ and await the next iteration (Line 19).
Finally, if the number of left unaligned source entities no longer decreases, we stop the process and output the repaired results (Line 21).}

\smallskip\noindent\textbf{\revise{Analysis.}}
\revise{
To ensure the correctness of Algorithm \ref{alg:one_to_many},
we check conflicts in two directions: from source to target KGs (Line 4) and from target to source KGs (Line 8), ensuring that no one-to-many conflicts exist in the output.
In addition, the resulting one-to-one alignment either has high similarity or has higher explanation confidence, contributing to more accurate results.
In this process, each alignment update will have an impact on the confidence of other alignment.
Therefore, it is difficult to prove its convergence.
Thus, we set a condition to ensure that the algorithm can be terminated (Line 21).
For time complexity, in the worst case, it is $O(|\mathcal{E}_1'|^2)$. 
However, in practice, the number of unaligned entities left after one-to-many conflict resolution (Line 1) is much smaller than that of total entities, i.e., $|\mathcal{E}_1'|<|\mathcal{E}_1|$. 
Besides, the loop will break (Line 9 and Line 18) if the algorithm finds a one-to-one alignment result, resulting in much less complexity on average in practice.
}

\begin{algorithm}[!tb]
\caption{EA repair for low-confidence conflicts}
\label{alg:low_confidence}
{\small
\KwIn{Seed EA \revise{$\mathcal{A}_{train}$}; Unaligned source entities \revise{$\mathcal{E}_1'$};  EA \revise{$\mathcal{A^*}$}; 
number of candidate entities \revise{\revise{$k$}}.}
\KwOut{Repaired EA $\mathcal{A}^*$.}
$lastLen \leftarrow -1$\;
\While {True}{
    $\revise{\mathcal{E}'_{1low}},\revise{\mathcal{A}_{low}} \leftarrow$ \texttt{LowConfidence}(\revise{$\mathcal{A}^*$})\;  
    $\revise{\mathcal{E}_1'} \leftarrow \revise{\mathcal{E}_1'}.\texttt{union}(\revise{\mathcal{E}'_{1low}}); \revise{\mathcal{A}^*} \leftarrow \revise{\mathcal{A}^*}.\texttt{diff}(\revise{\mathcal{A}_{low}})$\;
    \lIf{$lastLen>-1$ \textbf{and} $\texttt{\tt len}(\revise{\mathcal{E}_1'})$ $\geq$ $lastLen$}{\textbf{break}}
    \lElse{$lastLen \leftarrow \texttt{len}(\revise{\mathcal{E}_1')}$}
    $\revise{\mathcal{E}'_{1new}} \leftarrow \texttt{set}()$\;
    \For{$e_1 \in \revise{\mathcal{E}_1'}$}{
        $candidateList \leftarrow$ \texttt{Candidate}($e_1$, \revise{$\mathcal{A^*}$})\;
        $ScoreList \leftarrow []$\;
        \For{$e_c \in candidateList$}{
            $x_c \leftarrow$ \texttt{Exp}($e_1, e_c, \revise{\mathcal{A^*}},\revise{\mathcal{A}_{train}}$)\;
            $g_c \leftarrow$ \texttt{ADGConstruct}($x_c$)\;
            $score \leftarrow g_c.conf + \alpha \times \texttt{sim}(e_1,e_c)$\;
            $ScoreList.\texttt{append}(score)$\;
        }
        $SortedScoreList, Rank \leftarrow ScoreList.\texttt{sort}()$\;
        
        \For{$j$ $\leftarrow 0$ \KwTo $\texttt{\tt min}(\revise{k}, \texttt{\tt len}(Rank)) - 1$}{
            $e_2 \leftarrow candidateList[Rank[j]]$\;
            \If{$e_2$ is not aligned}
            {
                $\revise{\mathcal{A^*}}.\texttt{add}((e_1, e_2))$; \textbf{break}\;
            }
            \Else{
                Get the entity $e_1'$ aligned with $e_2$\;
                $x_2 \leftarrow$ \texttt{Exp}($e_1', e_2, \revise{\mathcal{A^*}},\revise{\mathcal{A}_{train}}$)\;
                $g_2 \leftarrow$ \texttt{ADGConstruct}($x_2$)\;
                $score_2 \leftarrow g_2.conf + \alpha \times \texttt{sim}(e_1',e_2)$\;
                \If{$SortedScoreList[j] > score_2$}
                {
                    $\revise{\mathcal{A^*}}.\texttt{add}((e_1, e_2)); \revise{\mathcal{A^*}}.\texttt{remove}((e_1', e_2))$;\\
                    $\revise{\mathcal{E}'_{1new}}.\texttt{add}(e_1')$; \textbf{break}\;
                }
            }
        }
        \lIf{$e_1$ is not aligned}{$\revise{\mathcal{E}'_{1new}}.\texttt{add}(e_1)$}
    }
    $\revise{\mathcal{E}_1'} \leftarrow \revise{\mathcal{E}'_{1new}}$\;
}}
\end{algorithm}

\subsection{\revise{Low-confidence Conflicts}}
These conflicts indicate some alignment results with low explanation confidence are considered to be potentially incorrect. 

\smallskip\noindent\textbf{\revise{Conflict detection.}}
We detect these conflicts after the resolution of one-to-many conflicts. 
As depicted in Fig.~\ref{fig:conflict1}(c), after resolving one-to-many conflicts, the matched neighbor entity pair (Washington, New York) of incorrect alignment (Idaho, State of Oregon) will be deleted.
This makes the explanation for this alignment no longer sufficient and leads to low confidence.
\revise{
If an EA pair does not have \emph{strongly-influential} edges in its ADG, its confidence will be low, like in the above example.
This can also be confirmed by the process of calculating confidence.
Therefore, we can judge whether the confidence is low based on whether there are \emph{strongly-influential} edges in the ADG.
Sometimes this condition is too strong, and it can be relaxed appropriately. 
We can set the low confidence threshold $\beta$ based on the threshold $\theta$, which is used for judging whether the confidence calculated by \emph{strongly-influential} edges is sufficient in Eq. \eqref{conf_compute}.
It can be regarded as a binary classification problem and thus we set $\theta=0$.
Then, we set $\beta=\texttt{sigmoid}(\theta)$.

}

\smallskip\noindent\textbf{\revise{Conflict repair.}}
The repair process is shown in Algorithm~\ref{alg:low_confidence}.
We first remove the alignment with low explanation confidence (Lines 3--4).
Next, there is a condition to determine whether to terminate the iteration (Line~5). 
We continue the repair process until the number of unaligned source entities no longer decreases. 
Given a conflict to repair, we search for entities that can form an explanation with higher confidence as candidate entities (Line 9).
We first find the target entities with aligned neighbors, because their confidence is guaranteed to be greater than 0.5.
We remove the target entities that present low confidence with the source entity.
Then, we generate the explanations and construct ADGs for the candidate entities and the source entity to be aligned (Lines 12--13).
In this repair process, we no longer rely solely on the confidence of the explanations to select better alignment,
because the explanations that we generate are localized.
Consequently, the low-confidence explanation may arise from the limitations of our method in incorporating global information.
To reduce the errors in repair, we introduce an additional score, named alignment score, which is a combination of entity embedding similarity provided by the original model $f$, and the confidence in ADGs (Line 14). 
By doing so, we seek a balance between local and global information to make more informed alignment decisions for resolving low-confidence conflicts.
Then, we sort the candidate entities according to their alignment scores with the source entity (Lines 16--18).
Likewise, if the selected candidate entity $e_2$ has not been aligned yet, we will align it with the current entity $e_1$ (Lines 19--20).
Otherwise, if $e_2$ already has an aligned entity $e_1'$, we will calculate its alignment score $score_2$ with $e_1'$ (Lines 22--25).
We compare $score_2$ with the alignment scores of $e_1$ and $e_2$ and select the one with the higher score as the correct alignment (Lines 26--28).
If $e_1$ cannot be aligned with entities at this stage, we add $e_1$ to $\mathcal{E}'_{1new}$ and await the next repair iteration (Line 29).
Finally, for a small number of source entities whose alignment still cannot be found after the above repair operations, we resort to greedy matching to obtain the alignment with entities from the target KG that have not been aligned.
Then, these results are combined with repaired alignment to get our final output $\mathcal{A}^*$.

\revise{\smallskip\noindent\textbf{Analysis.}}
\revise{
In Algorithm \ref{alg:low_confidence}, we can obtain all the low-confidence alignment filtered by a threshold, ensuring that no low-confidence conflicts exist in the output. 
In the repair process, we delete or replace unaligned source entities to guarantee a decrease in the number of unaligned entities.
It is also difficult to prove the process is convergent, so we set the condition in Line 5 to ensure the algorithm ends after iterations.
In the algorithm, the outer loop will be repeated at most $|\mathcal{E}_1'|$ times. 
But it can end earlier when the number of unaligned source entities no longer decreases.
In each loop, we get at most $|\mathcal{E}_1'|$ unaligned source entities. 
Therefore, the time complexity is $O(|\mathcal{E}_1'|^2)$ in the worst case.
However, similar to Algorithm~\ref{alg:one_to_many}, the number of unaligned entities is much smaller than that of total entities, i.e., $|\mathcal{E}_1'|<|\mathcal{E}_1|$. Moreover, the pruning strategies result in less time complexity on average.
}

\section{Experiments}\label{sect:exp}
In this section, we present detailed experiments on explanation generation (Section~\ref{exp:generation}) and EA repair (Section~\ref{exp:repair}).
\revise{In addition, we compare \framework with large language models (LLMs) (Section~\ref{exp:LLM}) on EA explanation generation and EA verification.
We also evaluate the robustness of \framework against EA noise (Section~\ref{exp:noise}).}
The source code is available at link.\footnote{\url{https://github.com/nju-websoft/ExEA}}

\subsection{Setup}

\subsubsection{Environment}
All our experiments, including the original training and evaluation of models, are conducted on a server equipped with two Intel Xeon Gold 5122 3.6GHz CPUs, two NVIDIA RTX A6000 GPUs, and Ubuntu 18.04 LTS.

\subsubsection{Datasets}
\revise{We evaluate \framework on three cross-lingual EA datasets (ZH-EN, JA-EN, and FR-EN) from DBP15K~\cite{JAPE} and two heterogeneous datasets (DBP-WD-V1 and DBP-YAGO-V1) from OpenEA~\cite{OpenEA}.}
Each dataset comprises 15 thousand EA pairs for training and evaluation.

\subsubsection{EA models}
The \framework framework is applicable to any embedding-based EA model.
To showcase its generality and flexibility, we conduct detailed experiments based on four representative models as follows, each belonging to two categories of methods based on either TransE~\cite{TransE} or GCN~\cite{GCN}.

\begin{itemize}
    \item \textbf{MTransE}~\cite{MTransE} is a pioneering EA method based on embeddings. 
    It extends TransE~\cite{TransE} for embedding learning over multiple KGs, where relations are interpreted as translations from head entities to tail entities.
    It minimizes the distance between aligned entities.
    
    \smallskip\item \textbf{AlignE}~\cite{BootEA} is another translation-based method that enhances the alignment task through its loss function. 
    It employs a robust negative sampling method to differentiate similar entities, thereby significantly improving the accuracy of the alignment learning process.
    
    \smallskip\item \textbf{GCN-Align}~\cite{GCNAlign} is the first method that incorporates GCNs for embedding-based EA. 
    It captures similar subgraphs to learn the similarity between entities.
    
    \smallskip\item \textbf{Dual-AMN}~\cite{DualAMN} is another method based on GCNs, leveraging proxy matching and hard negative sampling techniques. 
    Among recent EA methods that solely rely on structure information, it has demonstrated superior performance and stands out as the best performer.
\end{itemize}

\subsection{Explanation Generation Experiments}\label{exp:generation}
The goal is to assess the explanations generated by \framework and explore the challenges in EA explanation generation.

\begin{table*}[!t]
\centering
\caption{\revise{Explanation generation using first-order triples on DBP15K and OpenEA.}}
\label{tab:eg1}
\resizebox{\textwidth}{!}{
\setlength{\tabcolsep}{0.9em}
\begin{tabular}{|l|l|c|c|c|c|c|c|c|c|c|c|c|c|c|c|}
\hline
\multirow{2}{*}{EA models} & \multirow{2}{*}{Exp. methods} & \multicolumn{2}{|c}{ZH-EN} & \multicolumn{2}{|c|}{JA-EN} & \multicolumn{2}{c|}{FR-EN}& \multicolumn{2}{c|}{\revise{DBP-WD}}& \multicolumn{2}{c|}{\revise{DBP-YAGO}} \\
\cline{3-12}
& & Fidelity & Sparsity & Fidelity & Sparsity & Fidelity & Sparsity & \revise{Fidelity} & \revise{Sparsity} & \revise{Fidelity} & \revise{Sparsity} \\ 
\hline
& EALime & 0.676 & 0.547 & 0.705 & 0.517 & 0.551 & 0.574 & \revise{0.757} & \revise{0.539} & \revise{0.837} & \revise{0.370} \\ 
 & EAShapley & 0.715 & 0.547 & 0.875 & \textbf{0.518} & 0.745 & \textbf{0.575} & \revise{0.759} & \revise{0.540} & \revise{0.830} & \revise{0.370} \\ 
MTransE & \revise{Anchor} & \revise{0.676} & \revise{0.546} & \revise{0.746} & \revise{0.517} & \revise{0.721} & \revise{0.574} & \revise{0.722} & \revise{0.538} & \revise{0.827} & \revise{0.370} \\ 
& \revise{LORE} & \revise{0.687} & \revise{0.531} & \revise{0.739} & \revise{0.499} & \revise{0.720} & \revise{0.560} & \revise{0.722} & \revise{0.524} & \revise{0.824} & \revise{0.341} \\ 
& ExEA (ours) & \textbf{0.874} & \textbf{0.559} & \textbf{0.909} & 0.510 & \textbf{0.874} & 0.574  & \revise{\textbf{0.787}} & \revise{\textbf{0.533}} & \revise{\textbf{0.888}} & \revise{\textbf{0.371}} \\ 
\hline
&EALime & 0.433 &  0.534& 0.476 & 0.556 & 0.480 & 0.572 & \revise{0.763} & \revise{0.471} & \revise{0.717} & \revise{0.371} \\ 
& EAShapley & 0.658 & 0.534 & 0.698 & 0.555 & 0.728 & 0.570 & \revise{0.771} & \revise{0.471} & \revise{0.760} & \revise{\textbf{0.372}} \\ 
AlignE & \revise{Anchor} & \revise{0.437} & \revise{0.523} & \revise{0.575} & \revise{0.526} & \revise{0.539} & \revise{0.571} & \revise{0.747} & \revise{0.471} & \revise{0.763} & \revise{\textbf{0.372}} \\ 
& \revise{LORE} & \revise{0.467} & \revise{0.541} & \revise{0.520} & \revise{0.529} & \revise{0.497} & \revise{0.576} & \revise{0.729} & \revise{0.487} & \revise{0.776} & \revise{0.333} \\ 
&ExEA (ours) & \textbf{0.874} & \textbf{0.549} & \textbf{0.754} & \textbf{0.581} & \textbf{0.745} & \textbf{0.580} & \revise{\textbf{0.806}} & \revise{\textbf{0.499}} & \revise{\textbf{0.867}} & \revise{0.371} \\ 
\hline
& \revise{EALime} & \revise{0.182} & \revise{0.614} & \revise{0.212} & \revise{0.587} & \revise{0.238} & \revise{0.627} & \revise{0.533} & \revise{0.479} & \revise{0.695} & \revise{0.301} \\ 
& \revise{EAShapley} & \revise{0.330} & \revise{0.618} & \revise{0.314} & \revise{0.583} & \revise{0.346} & \revise{0.583} & \revise{0.554} & \revise{0.475} & \revise{0.673} & \revise{0.300} \\ 
GCN-Align& \revise{Anchor} & \revise{0.253} & \revise{\textbf{0.654}} & \revise{0.277} & \revise{0.559} & \revise{0.186} & \revise{0.587} & \revise{0.360} & \revise{0.443} & \revise{0.612} & \revise{0.323} \\ 
& \revise{LORE} & \revise{0.172} & \revise{0.634} & \revise{0.213} & \revise{0.610} & \revise{0.214} & \revise{0.641} & \revise{0.340} & \revise{0.492} & \revise{0.614} & \revise{0.332} \\ 
&ExEA (ours) & \textbf{0.876} & 0.638 & \textbf{0.873} & \textbf{0.628} & \textbf{0.870} & \textbf{0.645} & \revise{\textbf{0.984}} & \revise{\textbf{0.494}} & \revise{\textbf{0.990}} & \revise{\textbf{0.346}} \\ 
\hline
&EALime & 0.643 & 0.443 & 0.654 & 0.471 & 0.643 & 0.460 & \revise{0.823} & \revise{0.370} & \revise{0.840} & \revise{0.316} \\ 
 & EAShapley & 0.824 & 0.466 & 0.867 & 0.469 & 0.825 & 0.457 & \revise{0.840} & \revise{0.369} & \revise{0.852} & \revise{0.316} \\ 
Dual-AMN & \revise{Anchor} & \revise{0.805} & \revise{0.466} & \revise{0.876} & \revise{0.469} & \revise{0.819} & \revise{0.457} & \revise{0.877} & \revise{0.369} & \revise{0.865} & \revise{0.317} \\ 
& \revise{LORE} & \revise{0.808} & \revise{0.466} & \revise{0.871} & \revise{0.469} & \revise{0.809} & \revise{\textbf{0.485}} & \revise{0.890} & \revise{0.346} & \revise{0.865} & \revise{0.316} \\ 
&ExEA (ours) & \textbf{0.959} & \textbf{0.476} & \textbf{0.965} & \textbf{0.473} & \textbf{0.974} & 0.480 & \revise{\textbf{0.917}} & \revise{\textbf{0.384}} & \revise{\textbf{0.937}} & \revise{\textbf{0.330}} \\ 
\hline
\end{tabular}}
\end{table*}

\begin{table*}[!t]
\centering
\caption{\revise{Explanation generation using triples within second-order on DBP15K and OpenEA.}}
\label{tab:eg2}
\resizebox{\textwidth}{!}{
\setlength{\tabcolsep}{0.9em}
\begin{tabular}{|l|l|c|c|c|c|c|c|c|c|c|c|c|c|c|c|c|}
\hline
\multirow{2}{*}{EA models} & \multirow{2}{*}{Exp. methods} & \multicolumn{2}{|c}{ZH-EN} & \multicolumn{2}{|c|}{JA-EN} & \multicolumn{2}{c|}{FR-EN}& \multicolumn{2}{c|}{\revise{DBP-WD}} & \multicolumn{2}{c|}{\revise{DBP-YAGO}}  \\
\cline{3-12}
& & Fidelity & Sparsity & Fidelity & Sparsity & Fidelity & Sparsity & \revise{Fidelity} & \revise{Sparsity} & \revise{Fidelity} & \revise{Sparsity} \\ 
\hline
& EALime & 0.391 & 0.439 & 0.488 & 0.429 & 0.456 & 0.440 & \revise{0.451} & \revise{0.422} & \revise{0.564} & \revise{0.343} \\ 
& EAShapley & 0.449 & 0.441 & 0.502 & 0.438 & 0.477 & 0.434 & \revise{0.449} & \revise{0.421} & \revise{0.571} & \revise{0.343} \\ 
Dual-AMN & \revise{Anchor} & \revise{0.428} & \revise{0.428} & \revise{0.507} & \revise{0.432} & \revise{0.460} & \revise{0.440} & \revise{0.437} & \revise{0.419} & \revise{0.669} & \revise{0.343} \\ 
& \revise{LORE} & \revise{0.430} & \revise{0.428} & \revise{0.502} & \revise{0.433} & \revise{0.464} & \revise{0.440} & \revise{0.442} & \revise{0.421} & \revise{0.670} & \revise{0.342} \\ 
& ExEA (ours) & \textbf{0.921} & \textbf{0.442} & \textbf{0.929} & \textbf{0.438} & \textbf{0.937} & \textbf{0.441} & \revise{\textbf{0.939}} & \revise{\textbf{0.434}} & \revise{\textbf{0.939}} & \revise{\textbf{0.355}} \\ 
\hline
\end{tabular}}
\end{table*}

\subsubsection{Baselines}
To date, there have been no studies explaining the results of embedding-based EA models. 
Consequently, we transfer methods commonly used in other explanation generation scenarios to create \revise{four} baselines for the EA task.

\begin{itemize}
    \item \textbf{EALime} is a method inspired by LIME~\cite{LIME}. 
    \revise{In EALime, we consider a triple as a feature and leverage linear regression to perform the local operation of fitting the EA model.
    We generate perturbed data by sampling triples around the entity pair to be explained and utilize the similarity between the entity representations obtained under the perturbed data as the prediction for a linear regression model.}
    For certain GCN-based models, we can directly employ the trained model $f$ to learn the perturbed triples to obtain the perturbed results. 
    As for the TransE-based models, given that they only provide embedding representations of entities and relations, in order to obtain the output of the original model following a disturbance, we use the model $f$ to average all the representations that translate neighbor entities to the central entity through relations.
    For instance, given the perturbed triple set $\{(e_1,r_1,e_1'),(e_1,r_2,e_2'),\dots,(e_1,r_n,e_n')\}$ from one KG, where the central entity is $e_1$, the result is as follows:
    \begin{align}
    \mathbf{e}_1 = \frac{1}{n}\sum_{i=1}^n(\mathbf{e}_i' - \mathbf{r}_i).
    \end{align}
    
    The similarity kernel $\pi_x$ in the LIME framework is used to measure the similarity between the perturbed sample and the original sample.
    Here, we use the output $\mathbf{e}_1',\mathbf{e}_2'$ of the perturbed sample $\mathcal{T}'$ after passing through the model $f$ to compute similarity with the result embeddings $\mathbf{e}_1,\mathbf{e}_2$ of the original sample $\mathcal{T}$. It is calculated as follows:
    \begin{align}
    \pi_x(\mathcal{T}') = \frac{1}{2}\big({\rm sim}(\mathbf{e}_1',\mathbf{e}_1) + {\rm sim}(\mathbf{e}_2',\mathbf{e}_2)\big).
    \end{align}

    \smallskip\item \textbf{EAShapley} is a method based on the Shapley value~\cite{Shapley}, which implies that the model's prediction is the outcome of the cooperation among different features.
    Similar to EALime, we also consider a triple as a feature and employ the EA model $f$ as a value function to evaluate the combined feature.
    For the TransE-based methods, we follow EALime to obtain the results of the perturbed samples.
    Due to the vast number of triples, we employ an approximate method to handle the computations.
    When considering only first-order neighbors, we employ the Monte Carlo algorithm for approximate calculations to achieve relatively accurate results. 
    When considering second-order triples, the Monte Carlo algorithm is still time-consuming. Therefore, we employ the KernelSHAP-based method~\cite{SHAP}.
    This method is similar to EALime, with the exception that the similarity kernel is replaced by the Shapley kernel. 
    Here, $\pi_x$ is calculated as follows:
    \begin{align}
    \pi_x(\mathcal{T}') = \frac{|\mathcal{T}| - 1}{(|\mathcal{T}| \,\text{choose}\, |\mathcal{T}'|)|\mathcal{T}'|(|\mathcal{T}| - |\mathcal{T}'|)}.
    \end{align}
    \item \revise{\textbf{Anchor}~\cite{Anchor} explains predictions of classification models with rules.
    To adapt Anchor for EA,
    we model the EA task as a classification problem with a similarity threshold to classify entity pairs.
    If the similarity of two entity embeddings exceeds the threshold, the pair is classified as a positive alignment. Otherwise, it is a negative.}
    \smallskip\item \revise{\textbf{LORE}~\cite{lore} can obtain both decision rules and counterfactual rules for classification predictions.
    To adapt LORE for EA, we do the same data processing as above.}
\end{itemize}

\subsubsection{Evaluation methods and metrics}
To assess the quality of explanations, we use the fidelity metric.
The explanations that we define for the EA results are crucial for achieving accurate predictions. 
Therefore, it is essential to evaluate whether the explanations that we discover enable the model to maintain its prediction accuracy.
To measure this metric, we employ a method similar to \cite{kelpie} to sample 1,000 cases from the correct predictions by the EA model. 
For these 1,000 alignment cases, the accuracy rate is 100\%. 
Subsequently, we process each sample as follows: assuming the candidate triples provided for the current sample is $\mathcal{T}$, and the found explanation is $\mathcal{T}'$, we remove $\mathcal{T} - \mathcal{T}'$ from the dataset. 
Finally, we retrain the model on the processed dataset and measure how much of the 1,000 alignment are still predicted accurately. 
The ratio of correct predictions is used as the fidelity.
Furthermore, we aim for shorter explanations to be preferred, as a shorter explanation implies higher fidelity by capturing a higher density of useful information. 
In other words, the shorter the explanation, the better, as it can more effectively reflect the quality of the obtained information.
To achieve this objective, we introduce the sparsity metric for computation.
For candidate triples $\mathcal{T}$, explanation $\mathcal{T}'$, the sparsity is calculated as follows: 
\begin{align}
\text{sparsity}=1 - \frac{|\mathcal{T}'|}{|\mathcal{T}|}.
\end{align}

\framework does not require pre-selecting the explanation length. 
For a fair comparison between methods, we adjust the parameters of baseline methods to select the explanation length to ensure that the sparsity is as close as possible to that of \framework. 

\subsubsection{Main results}
The experiments involve using first-order triples and the triples within the second order as candidates for explanation search, respectively.
The main results are presented in Table \ref{tab:eg1} (first-order) and Table \ref{tab:eg2} (within the second-order). 
We mainly consider the first experiment for two reasons.
The first is for the readability of EA explanations.
When the range of candidate triples is relatively large, it becomes possible to encounter triples that are distant from the entity pairs being explained. 
This can make it challenging to comprehend the relationship between them and the alignment that requires explanations.
The second is for certain original models, like the TransE-based models~\cite{MTransE}, that solely utilize first-order triples.
Consequently, when using the baseline method, which relies on the original model to process perturbed data, it becomes limited to handling only first-order triples and cannot effectively process data involving higher-order triples. 
From the results presented in both tables, it is evident that \framework has achieved the best performance on all datasets and models when considering similar sparsity levels. 
This further supports our heuristic idea that the EA model's predictions are strongly related to the presence of similar matching parts around the entity.
It should be noted that GCN-Align itself does not distinguish the triple information related to neighbor entities, which consequently hinders the ability of the baseline methods that rely on the original model to identify which triples are crucially related to neighbors. 
As a result, the baseline methods can only extract all triples associated with important entities.
However, due to the limitation of sparsity, it may only obtain the triples of a few entities, resulting in a substantial loss of critical information and causing a significant decline in the performance of the baseline methods.
From the other results, it is evident that EAShapley performs as the second-best method, sometimes coming close to the results achieved by \framework, while EALime performs the worst. 
Based on this significant experimental finding, it can be concluded that in the local context, using a single triple as a feature cannot simply apply a linear model to fit the EA model. 
We believe that this outcome aligns with our heuristic idea, as the EA model requires matching triples in both KGs to work in synergy, and it cannot treat each triple as independent.
The nonlinear Shapley value calculation considers the interactions between triples, thereby effectively capturing similar triples between the two KGs.
This observation is reflected in the results, where most of the EAShapley results have matched triples.
We believe that this characteristic plays a significant role in enabling EAShapley to approximate \framework results in certain cases.
However, \revise{it is also evident from the results that the Shapley values of some unmatched triples are ranked relatively high,  resulting in reduced effectiveness.} 
This can be attributed to two reasons: 
First, the calculation relies on local data after sampling, making it challenging to distinguish highly similar triples in the absence of global information.
Second, the inaccuracy in the results after inputting sampled data into the original model can also contribute to this outcome.
For instance, in cases where the TransE-based model is the original model, obtaining the results through approximation may negatively impact EAShapley, which relies on the feedback from the original model.
\revise{For the same reason, Anchor and LORE sometimes obtain rules of poor quality and thus fail to meet sufficient conditions.}
In contrast, \framework utilizes the global alignment from the final model and operates independently from the framework of the original model, effectively addressing the limitations of \revise{baselines} and achieving the best performance.

Regarding the experiments involving second-order triples as candidate triples, we conduct these experiments exclusively on Dual-AMN. 
This decision is made based on factors discussed earlier. 
MTransE and AlignE can only handle first-order neighbors,
rendering the baseline methods ineffective. 
GCN-Align, due to its disregard for relation information, poses challenges for the baseline methods in selecting triples, making it less meaningful to compare the results with \framework.
From the results, it can be observed that the performance of \framework, although slightly decreased, remains notably high.
This indicates that our method exhibits strong robustness when the candidate scope is expanded.
EAShapley, at this time, relies on the framework of KernelSHAP to use the linear model for fitting, which leads to a significant decline in effectiveness.

\begin{figure}[!t]
	\centering
	\includegraphics[width=\linewidth]{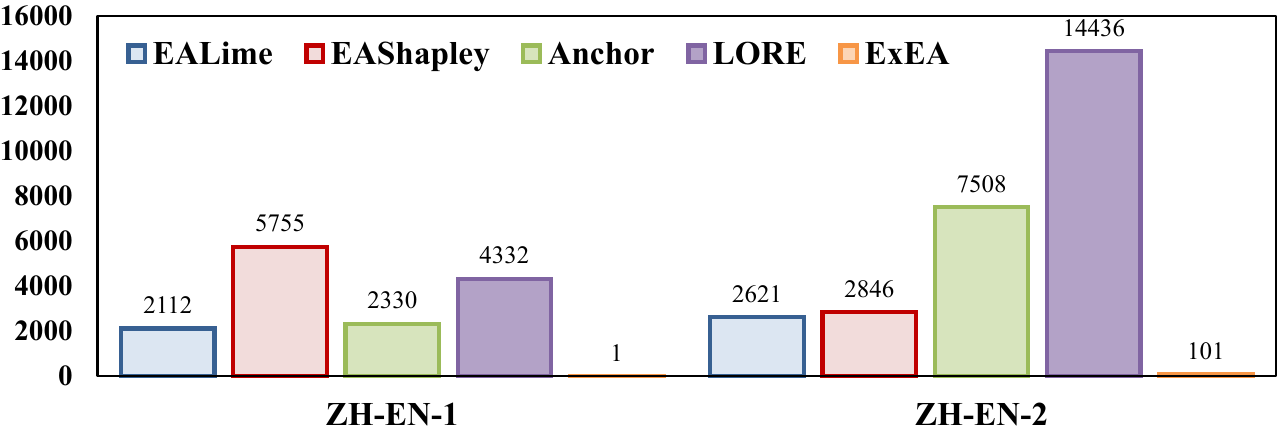}
	\caption{\revise{Time cost (s) of explanation generation for Dual-AMN on ZH-EN.}}
	\label{fig:time}
\end{figure}

\subsubsection{Comparison of efficiency}
As discussed earlier, existing methods for generating explanations usually involve traversing various feature combinations to determine the importance of different features. 
For instance, EALime and EAShapley both require obtaining different combinations of triples during the computation process.
Even when using sampling techniques for approximation, significant time consumption remains unavoidable. 
Moreover, since these methods rely on feedback from the original model, they further affect efficiency by adding to the time consumption during the model's output process.
Since we intend to use the explanations to repair the alignment later, we find that methods with high time consumption are not advisable.
Therefore, we conduct a time comparison of explanation generation methods for Dual-AMN on the ZH-EN dataset, including cases where candidate triples ranged from first-order triples to those that included second-order triples, i.e., (-1 vs -2 versions).
The results are shown in Fig.~\ref{fig:time}.
It is evident that \framework is significantly faster than other methods.
Moreover, EAShapley, which is based on KernelSHAP, demonstrates an improvement in time efficiency. 
Even when the candidate triples are in the second-order range, it is faster than when the candidate triples are in the first-order range.
This time efficiency improvement can be attributed to the use of the Monte Carlo algorithm in EAShapley, which needs to calculate the marginal effect of continuously adding triples to the current set in each simulation.
This process increases computation time but also enhances accuracy. 
\revise{LORE costs the most time due to the inefficiency of its genetic algorithm with multiple iterations.}

Overall, \framework stands out not only in the quality of the generated explanations but also in terms of time efficiency.
Consequently, we use \framework to efficiently perform explanation generation and subsequently repair alignment results.

\subsubsection{Case study}
\begin{figure}[!t]
\centering
\includegraphics[width=\linewidth]{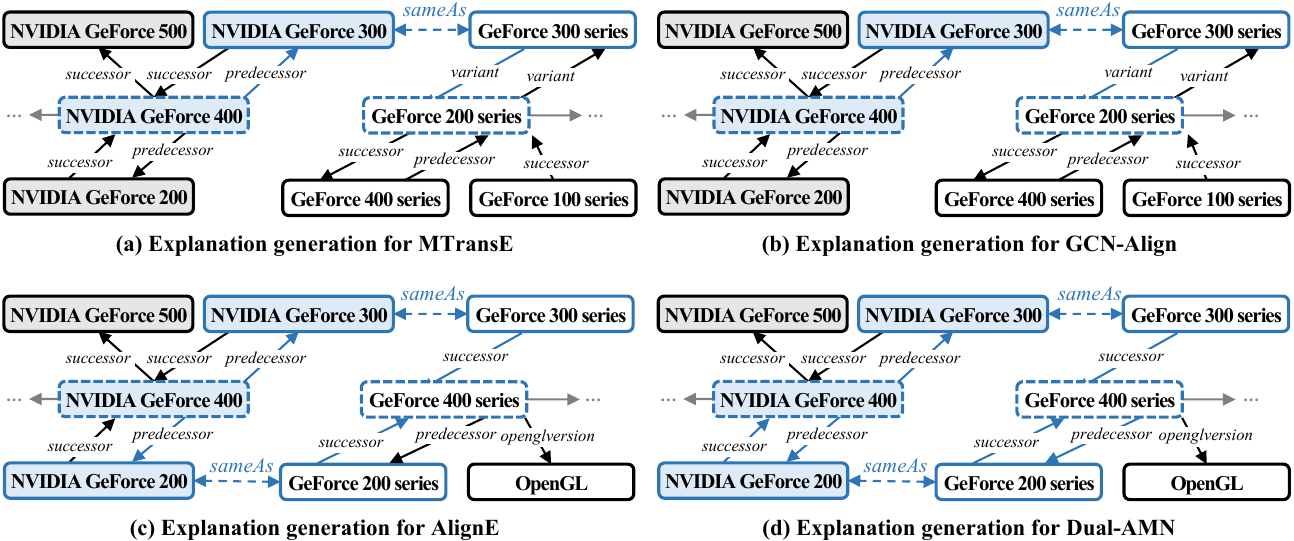}
\caption{\revise{Case study of our explanations for EA models. The two dotted boxes in each subgraph represent the predicted aligned entities, which may be incorrect. The blue boxes and arrows form our explanation for the prediction.}}
\label{fig:case}
\end{figure}

Fig.~\ref{fig:case} illustrates two examples of our explanations for the source entity ``NVIDIA GeForce 400'' and the possible counterparts found for it by different models. 
The blue part indicates our explanations.
From these explanations, we can not only intuitively understand why the model makes corresponding predictions, but also discern the characteristics and shortcomings of different models.
MTransE and GCN-Align are considered relatively simplistic models, as they fail to fully leverage the semantic information present in the data.
They just randomly align the entities adjacent to the known alignment entities ``NVIDIA GeForce 300'' and ``GeForce 300 series'' without effectively utilizing the relation semantics to distinguish between similar neighbor entities.
Consequently, this oversimplified process leads to incorrect alignment.
On the contrary, AlignE stands out by successfully learning the equivalence between the relation $predecessor$ and the reverse of the relation $successor$. 
It can effectively rectify misaligned entities from the aforementioned methods.
We believe the key factor behind this improvement lies in the implementation of strong negative sampling in AlignE, enabling it to adeptly discern the relation semantics among similar entities.
Compared with the explanations generated by AlignE, Dual-AMN offers a more comprehensive understanding, showcasing its proficiency in capturing the relational semantics in the data, and reflecting the powerful capabilities of its complex encoder.

\begin{table*}[!t]
\centering
\caption{\revise{EA repair results on DBP15K and OpenEA.}}
\label{tab:repair}
\resizebox{\textwidth}{!}{
\setlength{\tabcolsep}{0.8em}
\begin{tabular}{|l|c|c|c|c|c|c|c|c|c|c|c|c|c|c|c|c|c|}
\hline
\multirow{2}{*}{EA models} & \multicolumn{3}{|c}{ZH-EN} & \multicolumn{3}{|c|}{JA-EN} & \multicolumn{3}{c|}{FR-EN} & \multicolumn{3}{|c|}{\revise{DBP-WD}} & \multicolumn{3}{c|}{\revise{DBP-YAGO}}\\
\cline{2-16}
& Base & \framework & $\Delta \, acc$ & Base & \framework & $\Delta \, acc$ & Base & \framework & $\Delta \, acc$ & \revise{Base} & \revise{\framework} & \revise{$\Delta \, acc$} & \revise{Base} & \revise{\framework} & \revise{$\Delta \, acc$} \\ 
\hline
MTransE & 0.423 & 0.761 & +0.338 & 0.442 & 0.640 & +0.198 & 0.423 & 0.658 & +0.235 & \revise{0.404} & \revise{0.564} & \revise{+0.160} & \revise{0.569} & \revise{0.732} & \revise{+0.163} \\ 
AlignE & 0.488 & 0.705 & +0.217 & 0.460 & 0.676 & +0.216 & 0.483 & 0.724 & +0.241 & \revise{0.419} & \revise{0.602} & \revise{+0.183} & \revise{0.546} & \revise{0.723} & \revise{+0.177} \\ 
GCN-Align & 0.405 & 0.640 & +0.235 & 0.438 & 0.623 & +0.185 & 0.418 & 0.618 & +0.200 & \revise{0.421} & \revise{0.562} & \revise{+0.141} & \revise{0.594} & \revise{0.740} & \revise{+0.146} \\ 
Dual-AMN & 0.670 & 0.797 & +0.127 & 0.656 & 0.777 & +0.121 & 0.678 & 0.820 & +0.142 & \revise{0.621} & \revise{0.701} & \revise{+0.080} & \revise{0.730} & \revise{0.806} & \revise{+0.076} \\ 
\hline
\end{tabular}}
\end{table*}

\subsection{Entity Alignment Repair Experiments}\label{exp:repair}
In this section, we conduct experiments on repairing EA results based on the generated explanations and ADGs.

\subsubsection{Evaluation metric}
In this experiment, our primary focus is on EA performance. 
Therefore, the metric is accuracy, which represents the proportion of correctly aligned entity pairs.

\subsubsection{Main results}
Table \ref{tab:repair} presents the main results of EA repair on \revise{five} datasets with four models.
We find that \framework can effectively repair EA results and bring a significant improvement in alignment accuracy.
Some simple models, such as MTransE, after being repaired by \framework, can even approach or surpass the original performance of the state-of-the-art model Dual-AMN. 
For instance, MTransE, with our repair process, achieves an accuracy of 0.761 on the ZH-EN dataset, which outperforms the original Dual-AMN (0.670).
We also notice that the models based on TransE generally experience a greater improvement in performance compared to those based on GCNs. 
We attribute this observation to the similarity between the process of computing explanation confidence in \framework and the aggregation process in GCNs.
Consequently, during the repair process, some of the information provided may overlap with the existing knowledge of these models, leading to a relatively smaller improvement in performance.
On the other hand, the improvement in performance for Dual-AMN is notably smaller than that of other models. 
We believe this can be attributed to limitations in the dataset, which impose a certain upper limit on the model's accuracy in the EA task.
We observe instances where correct alignment becomes challenging, as there are even false target entities that have an equal or greater number of triples matching the source entities' triples compared to the correct target entities. 
This situation makes it difficult even for human evaluators to accurately accomplish the alignment.
We also see that the repair effectiveness is closely related to the differences in the datasets.
For instance, the JA-EN dataset is notably challenging. 
Even for some models that initially perform best on this dataset, the improvement after repair is minimal, such as MTransE and GCN-Align.
On the FR-EN dataset, the number of triples is noticeably higher than in other datasets, resulting in a higher data density. 
Therefore, compared to some simpler models, the improved versions, such as AlignE and Dual-AMN, can better utilize the data information. 
As a result, the improvement in performance on the FR-EN dataset is better than that on \revise{other datasets.
In datasets with heterogeneous schemata, the large semantic differences between relations pose a challenge to EA, leading to less substantial improvement compared to cross-lingual datasets in a similar schema.}

\begin{table}[!t]
\centering
\caption{{Ablation study on MTransE.}}
\label{tab:ablation}
\resizebox{\columnwidth}{!}{
\setlength{\tabcolsep}{0.9em}
\begin{tabular}{|l|c|c|c|c|c|c|c|c|c|c|c|c|c|}
\hline
 Methods & \multicolumn{1}{|c}{ZH-EN} & \multicolumn{1}{|c|}{JA-EN} & \multicolumn{1}{c|}{FR-EN} & \multicolumn{1}{|c|}{\revise{DBP-WD}} & \multicolumn{1}{|c|}{\revise{DBP-YAGO}}\\

\hline
\framework w/o $cr_1$& 0.750& 0.638 & 0.656 & \revise{0.563} & \revise{0.730} \\ 
\framework w/o $cr_2$& 0.515 & 0.486 & 0.458 & \revise{0.463} & \revise{0.636} \\ 
\framework w/o $cr_3$& 0.712 &0.605 &  0.619 & \revise{0.517} & \revise{0.678} \\ 
\framework & \textbf{0.761} & \textbf{0.640}& \textbf{0.658} & \revise{\textbf{0.564}} & \revise{\textbf{0.732}} \\ 
\hline
\end{tabular}}
\end{table}

\subsubsection{Ablation study}
To examine the effectiveness of different conflict resolution methods during the repair process, we conduct an ablation study on \framework using the MTransE model.
The results are shown in Table \ref{tab:ablation}.
$cr_1$ refers to the resolution for relation alignment conflicts. 
$cr_2$ refers to the resolution for one-to-many conflicts. 
$cr_3$ refers to the resolution for low-confidence conflicts.
From the results, it is evident that each conflict resolution method is effective.
The most significant improvement in performance is achieved by resolving one-to-many conflicts.
This is consistent with intuition because ensuring one-to-one alignment reduces a lot of erroneous alignment caused by similar entities, thus making the repair process more effective.
Resolving low-confidence conflicts also shows significant effectiveness on all datasets, showing that 
low confidence indicates erroneous alignment in most cases.
The improvement from resolving relation alignment conflicts is not very evident on some datasets, which is also understandable.
We believe this is related to the completeness of different KGs.
When resolving these conflicts, we need to obtain the $\neg sameAs$ rules based on the relation pairs that do not appear simultaneously between a pair of entities in the KG. 
If a KG has higher completeness, there will be fewer cases where the identified relation pairs are missing due to KG incompleteness.
As a result, the quality of obtained rules will be higher, contributing to higher accuracy in conflict resolution.

\begin{figure}[!t]
\centering
\includegraphics[width=\columnwidth]{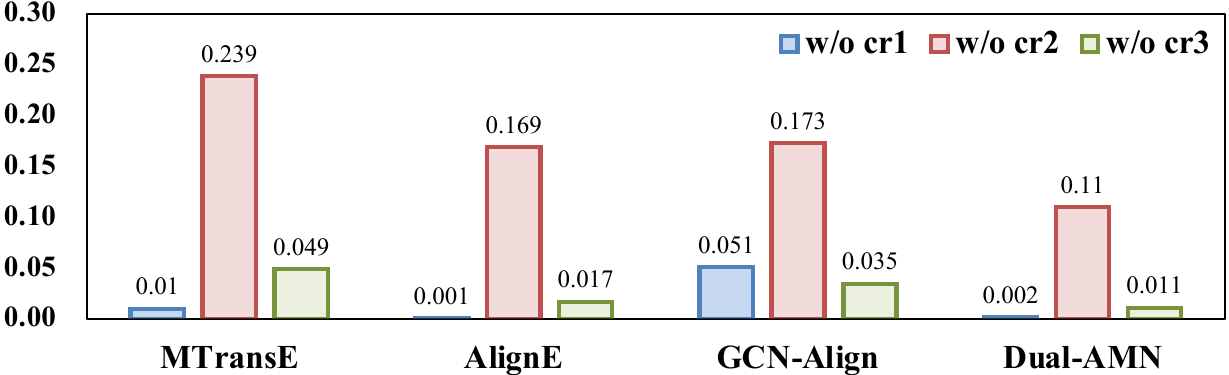}
\caption{The variation in repair effects across different models on ZH-EN.}
\label{fig:variation}
\end{figure}

\subsubsection{Variation in repair effects for different models}
We perform ablation on different conflict resolution methods for various models on the ZH-EN dataset.
The results are shown in Fig.~\ref{fig:variation}.
The values in the figure represent the decrease in accuracy after removing the corresponding component.
We can observe that AlignE and Dual-AMN, due to the use of negative sampling, have the ability to distinguish similar entities, resulting in relatively less improvement when resolving one-to-many conflicts.
It can also be observed that when the original performance is poor, resolving low-confidence conflicts leads to more significant improvements.
This is because such conflicts are often caused by unreliable predictions, which result in false matches around entity pairs. 
These false matches will provide low confidence or be eliminated after resolving one-to-many conflicts, leading to decreased confidence in the explanation of relevant EA pairs. 
After resolving relation alignment conflicts, GCN-Align shows the most significant improvement.
GCN-Align does not utilize relation information, so this operation essentially identifies additional important relation semantics, leading to a noticeable improvement.
For Dual-AMN, due to its complex and powerful relation modeling, the relevant relation semantics have already been learned effectively.
As a result, its improvement after resolving relation alignment conflicts is not as noticeable as compared to GCN-Align.
AlignE obtains the least improvement from resolving these conflicts. 
We believe this is because the resolution of such conflicts is actually consistent with the problem that AlignE aims to solve through negative sampling.
This conflict arises from the fact that there is a relation semantic structure between the aligned entities and their neighbors that does not support their alignment.
These entities are typically aligned based on their high similarity. 
Therefore, when resolving such conflicts, the corresponding alignment is treated as an error to distinguish similar entities. This aligns with the idea of negative sampling in AlignE.

\subsection{\revise{Experiments with Large Language Models}}\label{exp:LLM}
\revise{We hereby compare \framework with recent LLMs in terms of explanation generation and EA verification.}

\subsubsection{\revise{Explanation generation}}
\revise{
We implement two LLM-based explanation baselines using ChatGPT (GPT-3.5 Turbo).
The first method, called ChatGPT (perturb), borrows the idea in~\cite{llm_per}. 
It perturbs triples of an EA pair and gets the new prediction from the EA model for constructing prompts to ask ChatGPT.
The second method, called ChatGPT (match), follows the idea of our framework that EA explanations should consider matched triples around two entities.
It uses ChatGPT to find matched triples.
In this experiment, we randomly select 100 EA pairs from the test data of ZH-EN and DBP-WD, respectively.
The base EA models are MTransE and Dual-AMN.
Our framework uses first-order triples for explanation generation.
The results are presented in Table \ref{tab:eg_llm}.
We can see that our \framework still exhibits good performance.
ChatGPT (match), sharing similar principles with \framework, also achieves impressive results. 
This experiment further demonstrates the effectiveness of our key idea for EA explanation.
For ChatGPT (match), due to the hallucination issue in ChatGPT, it gets some incorrect answers during the triple matching process, causing performance to decrease.
Moreover, ChatGPT (match) does not incorporate model-relevant information. 
It has the potential to introduce matches that are unrelated to model predictions, which results in reduced sparsity. 
Some of these matches are even noisy and thus consequently diminish the overall effectiveness of ChatGPT (match).
ChatGPT (perturb) fails to perform well. 
Like other baselines in Section~\ref{exp:generation}, ChatGPT (perturb) is constrained by a limited local perspective and less accurate perturbation predictions. 
Additionally, the restricted input length of ChatGPT leads to the performance degradation of ChatGPT (perturb).
}

\begin{table}[!t]
\centering
\caption{\revise{Comparison with LLMs on explanation generation.}}
\label{tab:eg_llm}
\resizebox{\columnwidth}{!}{
\setlength{\tabcolsep}{0.6em}
\begin{tabular}{|l|l|c|c|c|c|c|c|c|c|c|c|c|c|c|c|c||c|c|}
\hline
\multirow{2}{*}{\revise{EA models}} & \multirow{2}{*}{\revise{Exp. methods}} & \multicolumn{2}{|c}{\revise{ZH-EN}}& \multicolumn{2}{|c|}{\revise{DBP-WD}} \\
\cline{3-6}
& & \revise{Fidelity} & \revise{Sparsity} & \revise{Fidelity} & \revise{Sparsity}\\ 
\hline
& \revise{ChatGPT (perturb)} & \revise{0.470} & \revise{0.605} & \revise{0.430} & \revise{0.643} \\ 
\revise{MTransE} & \revise{ChatGPT (match)} & \revise{\textbf{0.690}} & \revise{0.602}  & \revise{0.790} & \revise{0.671} \\ 
& \revise{ExEA} & \revise{\textbf{0.690}} & \revise{\textbf{0.629}} & \revise{\textbf{0.810}} & \revise{\textbf{0.689}} \\ 
\hline
& \revise{ChatGPT (perturb)} & \revise{0.430} & \revise{0.645} & \revise{0.480} & \revise{0.638} \\ 
\revise{Dual-AMN} & \revise{ChatGPT (match)} & \revise{0.780} & \revise{\textbf{0.647}} & \revise{0.810} & \revise{0.668} \\ 
& \revise{ExEA} & \revise{\textbf{0.820}} & \revise{\textbf{0.647}} & \revise{\textbf{0.910}} & \revise{\textbf{0.682}} \\ 
\hline
\end{tabular}}
\end{table}

\subsubsection{\revise{Entity alignment verification}}
\revise{
We further conduct experiments on ZH-EN and DBP-WD to compare the effectiveness of \framework and LLMs in EA verification.
We randomly choose 500 correct EA pairs and 500 incorrect EA pairs from the results obtained by MTransE and Dual-AMN, respectively.
We design a baseline inspired by~\cite{self-checker} that uses ChatGPT as a policy agent.
Specifically, each EA pair is treated as a claim, and the relation triples of the two aligned entities are considered as evidence.
The baseline uses ChatGPT to determine the correctness of the claim.
We report the precision, recall and F1-score results in Table~\ref{tab:av}.
We can observe that \framework still outperforms ChatGPT, and the slightly lower performance of ChatGPT can be attributed to two main reasons:
First, ChatGPT may not possess comprehensive information regarding certain entities, such as ``Charlie Soong'', and the evidence presented through local triples may be insufficient.
Second, we find that ChatGPT is insensitive to the numbers appearing in entity names, such as the names of NVIDIA GeForce GPUs in the case study of Fig.~\ref{fig:case}.
This issue causes the alignment of entities about different versions or generations to be incorrectly regarded as correct.
ChatGPT uses entity names, while our \framework uses graph structures. 
The two methods may be complementary to each other.
Thus, we merge the results of ChatGPT and \framework. 
We can see that the overall performance has significantly improved in comparison to their individual results. 
This demonstrates the complementary nature of structural information and textual knowledge in LLMs, indicating a promising direction for future exploration. 
The fusion of these two features holds the potential for enhancing EA performance and helping EA explanations.
}

\begin{table}[!t]
\centering
\caption{\revise{Comparison with LLMs on EA verification.}}
\label{tab:av}
\resizebox{\columnwidth}{!}{\Large
\setlength{\tabcolsep}{0.5em}
\begin{tabular}{|l|l|c|c|c|c|c|c|c|c|c|c|c|c|c|c|c|}
\hline
\multirow{2}{*}{\revise{EA models}} & \multirow{2}{*}{\revise{Ver. methods}} & \multicolumn{3}{|c}{\revise{ZH-EN}} &  \multicolumn{3}{|c|}{\revise{DBP-WD}}\\
\cline{3-8}
& & \revise{Prec.} & \revise{Recall} & \revise{F1} & \revise{Prec.} & \revise{Recall} & \revise{F1} \\ 
\hline
& \revise{ChatGPT} & \revise{0.823} & \revise{0.862} & \revise{0.842} & \revise{0.841} & \revise{0.970} & \revise{0.901} \\ 
\revise{MTransE} & \revise{\framework} & \revise{0.918} & \revise{0.938} & \revise{0.928} & \revise{0.846} & \revise{0.966} & \revise{0.902} \\ 
& \revise{ChatGPT + \framework} & \revise{\textbf{0.982}} & \revise{\textbf{0.986}} & \revise{\textbf{0.984}} & \revise{\textbf{0.960}} & \revise{\textbf{0.996}} & \revise{\textbf{0.977}} \\ 
\hline
& \revise{ChatGPT} & \revise{0.816} & \revise{0.826} & \revise{0.821} & \revise{0.815} & \revise{0.944} & \revise{0.875} \\ 
\revise{Dual-AMN} & \revise{\framework} & \revise{0.879} & \revise{0.940} & \revise{0.905} & \revise{0.911} & \revise{0.978} & \revise{0.943} \\ 
& \revise{ChatGPT + \framework} & \revise{\textbf{0.970}} & \revise{\textbf{0.984}} & \revise{\textbf{0.977}} & \revise{\textbf{0.967}} & \revise{\textbf{0.996}} & \revise{\textbf{0.981}} \\ 
\hline
\end{tabular}}
\end{table}

\subsection{\revise{Experiments of Entity Alignment with Noise}}\label{exp:noise}
\revise{Considering that real-life KGs are not always clean~\cite{kg_quality} and there may exist errors in seed EA,
we hereby conduct an experiment with noise to evaluate the robustness of our framework.
We add noise to the seed EA set (4,500 EA pairs in total) by randomly disrupting the entities in its 750 EA pairs.
We report the explanation generation results for MTransE and Dual-AMN on ZH-EN and DBP-WD in Table~\ref{tab:eg_noise}.
We can see that \framework remains the best, which aligns with our expectations, as explanation generation seeks to adhere to the model's predictions and is independent of the data noise.
The EA repair results are given in Table \ref{tab:repair_noise}. 
We can observe that although EA noise results in a decrease in the EA performance of MTransE and Dual-AMN, our \framework can still significantly improve their results, showing good robustness.}

\begin{table}[!t]
\centering
\caption{\revise{Explanation generation of EA with Noise.}}
\label{tab:eg_noise}
\resizebox{\columnwidth}{!}{
\setlength{\tabcolsep}{0.8em}
\begin{tabular}{|l|l|c|c|c|c|c|c|c|c|c|c|c|c|c|c|c||c|c|}
\hline
\multirow{2}{*}{\revise{EA models}} & \multirow{2}{*}{\revise{Exp. methods}} & \multicolumn{2}{|c}{\revise{ZH-EN (Noise)}}& \multicolumn{2}{|c|}{\revise{DBP-WD (Noise)}} \\
\cline{3-6}
& & \revise{Fidelity} & \revise{Sparsity} & \revise{Fidelity} & \revise{Sparsity} \\ 
\hline
& \revise{EALime} & \revise{0.642} & \revise{0.648} & \revise{0.681} & \revise{0.629} \\ 
 & \revise{EAShapley} & \revise{0.661} & \revise{0.648}  & \revise{0.679} & \revise{0.629}  \\
\revise{MTransE} & \revise{Anchor} & \revise{0.654} & \revise{0.647}  & \revise{0.616} & \revise{0.628}  \\
  & \revise{LORE} & \revise{0.623} & \revise{0.621}  & \revise{0.615} & \revise{0.643}  \\
& \revise{\framework} & \revise{\textbf{0.746}} & \revise{\textbf{0.656}} & \revise{\textbf{0.930}} & \revise{\textbf{0.672}}\\ 
\hline
& \revise{EALime} & \revise{0.509} & \revise{0.752} & \revise{0.576} & \revise{0.597} \\ 
 & \revise{EAShapley} & \revise{0.459} & \revise{0.753}  & \revise{0.534} & \revise{0.596}  \\
\revise{Dual-AMN}& \revise{Anchor} & \revise{0.484} & \revise{0.751}  & \revise{0.559} & \revise{0.596}  \\
  & \revise{LORE} & \revise{0.450} & \revise{0.754}  & \revise{0.384} & \revise{0.589}  \\
& \revise{\framework} & \revise{\textbf{0.910}} & \revise{\textbf{0.781}} & \revise{\textbf{0.785}} & \revise{\textbf{0.636}}\\ 
\hline
\end{tabular}}
\end{table}

\begin{table}[!t]
\centering
\caption{\revise{EA repair results of EA with Noise.}}
\label{tab:repair_noise}
\resizebox{\columnwidth}{!}{
\setlength{\tabcolsep}{0.9em}
\begin{tabular}{|l|c|c|c|c|c|c|c|c|c|c|c|c|c|c|c|c|c|}
\hline
\multirow{2}{*}{\revise{EA models}} & \multicolumn{3}{|c}{\revise{ZH-EN (Noise)}} & \multicolumn{3}{|c|}{\revise{DBP-WD (Noise)}}\\
\cline{2-7}
& \revise{Base} & \revise{\framework} & \revise{$\Delta \, acc$} & \revise{Base} & \revise{\framework} & \revise{$\Delta \, acc$}\\ 
\hline
\revise{MTransE} & \revise{0.422} & \revise{0.650} & \revise{+0.228} & \revise{0.404} & \revise{0.560} & \revise{+0.156}\\ 
\hline
\revise{Dual-AMN} & \revise{0.520} & \revise{0.694} & \revise{+0.174} & \revise{0.517} & \revise{0.627} & \revise{+0.110}\\ 
\hline
\end{tabular}}
\end{table}

\section{Related Work}\label{sect:related_work}

\subsection{Embedding-based Entity Alignment}
In recent years, embedding-based methods have drawn significant attention for EA between KGs, primarily because of their capability to effectively capture semantics and similarities in vector space.
In the embedding learning phase, existing models use either TransE variants \cite{TransE,MTransE,JAPE,BootEA} or GCN variants \cite{GCNAlign,RelationalEA,CUEA,DualAMN}.
In the alignment inference phase, most existing models use the greedy search while a few other models adopt bi-directional $k$NN search~\cite{MRAEA}, reinforcement learning~\cite{icde_ea_rl,tois_ea_rl}, and holistic matching~\cite{BootEA,AssignmentEA} to improve performance.
In this paper, we focus on the EA models that learn from KG structures with relation triples and do not consider those that additionally use side features (e.g., textual names or descriptions).
Interested readers can refer to recent surveys~\cite{OpenEA,tkde_ea,aiopen_ea,vldbj_ea} for more details.

\subsection{Explanations for Machine Learning}
Two widely-used methods for machine learning explanations are LIME~\cite{LIME} and Shapley values~\cite{Shapley}.
LIME-based methods~\cite{LIME,reliable} primarily concentrate on local model prediction explanations.
They assume that the model can be approximated as an interpretable model locally, often a linear model.
The typical approach involves perturbing an instance and fitting the perturbed data using an interpretable model that can elucidate the behavior of the original model under that particular instance.
In this way, it provides insights into how the model's predictions are influenced by individual data points.
Shapley value-based methods derive their inspiration from cooperative game theory, specifically Shapley values.
These methods aim to fairly distribute the model's prediction contribution among the features of an instance. 
Shapley values consider the interactions between features and all possible feature combinations to determine the feature importance.
SHAP \cite{SHAP} is a widely used method based on Shapley values. 
Its variant KernelSHAP combines the strengths of both LIME and Shapley values. 
By leveraging kernel approximation, KernelSHAP achieves a balance between computational efficiency and accuracy when estimating Shapley values.
The explanation method for link prediction \cite{kelpie} cannot be directly transferred to the EA task, because it requires perturbing training data to observe important features as explanations, whereas our main focus is on important triples surrounding two entities.
In addition, the currently popular methods \cite{gnnexplainer,xsubgraph,gstarx,PGM,Counter} used for explaining GCNs are not easily transferable to the EA task for several reasons.
First, some EA models \cite{BootEA,MTransE} are based on TransE, which operates differently from GCNs.
Second, many EA models are highly complex, such as Dual-AMN, involving intricate considerations of relations, which makes it challenging to straightforwardly apply GCN-specific explanation techniques.
\revise{In addition to these feature-based explanation methods, there are also rule-based generative explanation methods~\cite{Anchor,lore}.
These methods have the capability to generate sufficient conditions~\cite{Anchor} and even counterfactual conditions~\cite{lore} for guiding decision-making. 
They not only provide a more intuitive form of explanations but also enhance the overall performance.}
Therefore, in our paper, we choose the methods based on \revise{LIME, Shapley values, and rules} as baselines.

\section{Conclusion and Future Work}\label{sect:concl}
In this paper, we present the first work, \framework, which can generate explanations for the output of embedding-based EA models. 
Given two entities predicted as an alignment, \framework first matches their neighborhood subgraphs and then builds an alignment dependency graph with confidence to understand this alignment. 
After explanation generation, \framework further reasons over all the alignment dependency graphs to repair EA results by resolving alignment conflicts.
Experiments on benchmark datasets demonstrate the effectiveness and efficiency of \framework in EA explanation generation and result repair.
For future work, we plan to take the side features of entities into consideration.

\section*{Acknowledgments}
This work was supported by the National Natural Science Foundation of China (No. 62272219) and the Collaborative Innovation Center of Novel Software Technology and Industrialization.

\balance
\bibliographystyle{IEEEtran}
\bibliography{ref.bib}

\begin{thebibliography}{10}
\providecommand{\url}[1]{#1}
\csname url@samestyle\endcsname
\providecommand{\newblock}{\relax}
\providecommand{\bibinfo}[2]{#2}
\providecommand{\BIBentrySTDinterwordspacing}{\spaceskip=0pt\relax}
\providecommand{\BIBentryALTinterwordstretchfactor}{4}
\providecommand{\BIBentryALTinterwordspacing}{\spaceskip=\fontdimen2\font plus
\BIBentryALTinterwordstretchfactor\fontdimen3\font minus
  \fontdimen4\font\relax}
\providecommand{\BIBforeignlanguage}[2]{{%
\expandafter\ifx\csname l@#1\endcsname\relax
\typeout{** WARNING: IEEEtran.bst: No hyphenation pattern has been}%
\typeout{** loaded for the language `#1'. Using the pattern for}%
\typeout{** the default language instead.}%
\else
\language=\csname l@#1\endcsname
\fi
#2}}
\providecommand{\BIBdecl}{\relax}
\BIBdecl

\bibitem{sf}
S.~Melnik, H.~Garcia{-}Molina, and E.~Rahm, ``Similarity flooding: {A}
  versatile graph matching algorithm and its application to schema matching,''
  in \emph{ICDE}, 2002, pp. 117--128.

\bibitem{PARIS}
F.~M. Suchanek, S.~Abiteboul, and P.~Senellart, ``{PARIS}: Probabilistic
  alignment of relations, instances, and schema,'' \emph{Proc. {VLDB} Endow.},
  vol.~5, no.~3, pp. 157--168, 2011.

\bibitem{MTransE}
M.~Chen, Y.~Tian, M.~Yang, and C.~Zaniolo, ``Multilingual knowledge graph
  embeddings for cross-lingual knowledge alignment,'' in \emph{IJCAI}, 2017,
  pp. 1511--1517.

\bibitem{TransE}
A.~Bordes, N.~Usunier, A.~Garc{\'{\i}}a{-}Dur{\'{a}}n, J.~Weston, and
  O.~Yakhnenko, ``Translating embeddings for modeling multi-relational data,''
  in \emph{NIPS}, 2013, pp. 2787--2795.

\bibitem{GCN}
T.~N. Kipf and M.~Welling, ``Semi-supervised classification with graph
  convolutional networks,'' in \emph{ICLR}, 2017.

\bibitem{tkde_ea}
X.~Zhao, W.~Zeng, J.~Tang, W.~Wang, and F.~M. Suchanek, ``An experimental study
  of state-of-the-art entity alignment approaches,'' \emph{IEEE Trans. Knowl.
  Data Eng.}, vol.~34, no.~6, pp. 2610--2625, 2022.

\bibitem{vldbj_ea}
R.~Zhang, B.~D. Trisedya, M.~Li, Y.~Jiang, and J.~Qi, ``A benchmark and
  comprehensive survey on knowledge graph entity alignment via representation
  learning,'' \emph{VLDB J.}, vol.~31, no.~5, pp. 1143--1168, 2022.

\bibitem{RelationalEA}
X.~Mao, W.~Wang, H.~Xu, Y.~Wu, and M.~Lan, ``Relational reflection entity
  alignment,'' in \emph{CIKM}, 2020, pp. 1095--1104.

\bibitem{AliNet}
Z.~Sun, C.~Wang, W.~Hu, M.~Chen, J.~Dai, W.~Zhang, and Y.~Qu, ``Knowledge graph
  alignment network with gated multi-hop neighborhood aggregation,'' in
  \emph{AAAI}, 2020, pp. 222--229.

\bibitem{DualAMN}
X.~Mao, W.~Wang, Y.~Wu, and M.~Lan, ``Boosting the speed of entity alignment
  10{\texttimes}: Dual attention matching network with normalized hard sample
  mining,'' in \emph{WWW}, 2021, pp. 821--832.

\bibitem{MRAEA}
X.~Mao, W.~Wang, H.~Xu, M.~Lan, and Y.~Wu, ``{MRAEA}: An efficient and robust
  entity alignment approach for cross-lingual knowledge graph,'' in
  \emph{WSDM}, 2020, pp. 420--428.

\bibitem{icde_ea_rl}
W.~Zeng, X.~Zhao, J.~Tang, and X.~Lin, ``Collective entity alignment via
  adaptive features,'' in \emph{ICDE}, 2020, pp. 1870--1873.

\bibitem{tois_ea_rl}
W.~Zeng, X.~Zhao, J.~Tang, X.~Lin, and P.~Groth, ``Reinforcement learning-based
  collective entity alignment with adaptive features,'' \emph{ACM Trans. Inf.
  Syst.}, vol.~39, no.~3, pp. 26:1--26:31, 2021.

\bibitem{BootEA}
Z.~Sun, W.~Hu, Q.~Zhang, and Y.~Qu, ``Bootstrapping entity alignment with
  knowledge graph embedding,'' in \emph{IJCAI}, 2018, pp. 4396--4402.

\bibitem{OpenEA}
Z.~Sun, Q.~Zhang, W.~Hu, C.~Wang, M.~Chen, F.~Akrami, and C.~Li, ``A
  benchmarking study of embedding-based entity alignment for knowledge
  graphs,'' \emph{Proc. VLDB Endow.}, vol.~13, no.~11, pp. 2326--2340, 2020.

\bibitem{LIME}
M.~T. Ribeiro, S.~Singh, and C.~Guestrin, ````{Why} should {I} trust you?'':
  Explaining the predictions of any classifier,'' in \emph{KDD}, 2016, pp.
  1135--1144.

\bibitem{JAPE}
Z.~Sun, W.~Hu, and C.~Li, ``Cross-lingual entity alignment via joint
  attribute-preserving embedding,'' in \emph{ISWC}, 2017, pp. 628--644.

\bibitem{explainability}
H.~Yuan, H.~Yu, S.~Gui, and S.~Ji, ``Explainability in graph neural networks: A
  taxonomic survey,'' \emph{IEEE Trans. Pattern Anal. Mach. Intell.}, vol.~45,
  pp. 5782--5799, 2022.

\bibitem{Shapley}
B.~Rozemberczki, L.~Watson, P.~Bayer, H.~Yang, O.~Kiss, S.~Nilsson, and
  R.~Sarkar, ``The {Shapley} value in machine learning,'' in \emph{IJCAI},
  2022, pp. 5572--5579.

\bibitem{GCNAlign}
Z.~Wang, Q.~Lv, X.~Lan, and Y.~Zhang, ``Cross-lingual knowledge graph alignment
  via graph convolutional networks,'' in \emph{EMNLP}, 2018, pp. 349--357.

\bibitem{bert}
J.~Devlin, M.~Chang, K.~Lee, and K.~Toutanova, ``{BERT:} pre-training of deep
  bidirectional transformers for language understanding,'' in
  \emph{{NAACL-HLT}}, 2019, pp. 4171--4186.

\bibitem{SHAP}
S.~M. Lundberg and S.-I. Lee, ``A unified approach to interpreting model
  predictions,'' \emph{NeurIPS}, 2017.

\bibitem{Anchor}
M.~T. Ribeiro, S.~Singh, and C.~Guestrin, ``Anchors: High-precision
  model-agnostic explanations,'' in \emph{AAAI}, vol.~32, 2018.

\bibitem{lore}
R.~Guidotti, A.~Monreale, S.~Ruggieri, D.~Pedreschi, F.~Turini, and
  F.~Giannotti, ``Local rule-based explanations of black box decision
  systems,'' \emph{arXiv}, 2018.

\bibitem{kelpie}
A.~Rossi, D.~Firmani, P.~Merialdo, and T.~Teofili, ``Explaining link prediction
  systems based on knowledge graph embeddings,'' in \emph{SIGMOD}, 2022, pp.
  2062--2075.

\bibitem{llm_per}
N.~Kroeger, D.~Ley, S.~Krishna, C.~Agarwal, and H.~Lakkaraju, ``Are large
  language models post hoc explainers?'' \emph{arXiv}, 2023.

\bibitem{self-checker}
M.~Li, B.~Peng, and Z.~Zhang, ``Self-checker: Plug-and-play modules for
  fact-checking with large language models,'' \emph{arXiv}, 2023.

\bibitem{kg_quality}
B.~Xue and L.~Zou, ``Knowledge graph quality management: {A} comprehensive
  survey,'' \emph{IEEE Trans. Knowl. Data Eng.}, vol.~35, no.~5, pp.
  4969--4988, 2023.

\bibitem{CUEA}
X.~Zhao, W.~Zeng, J.~Tang, X.~Li, M.~Luo, and Q.~Zheng, ``Toward entity
  alignment in the open world: An unsupervised approach with confidence
  modeling,'' \emph{Data Sci. Eng.}, vol.~7, no.~1, pp. 16--29, 2022.

\bibitem{AssignmentEA}
X.~Mao, W.~Wang, Y.~Wu, and M.~Lan, ``From alignment to assignment:
  Frustratingly simple unsupervised entity alignment,'' in \emph{EMNLP}, 2021,
  pp. 2843--2853.

\bibitem{aiopen_ea}
K.~Zeng, C.~Li, L.~Hou, J.~Li, and L.~Feng, ``A comprehensive survey of entity
  alignment for knowledge graphs,'' \emph{AI Open}, vol.~2, pp. 1--13, 2021.

\bibitem{reliable}
D.~Slack, A.~Hilgard, S.~Singh, and H.~Lakkaraju, ``Reliable post hoc
  explanations: Modeling uncertainty in explainability,'' in \emph{NeurIPS},
  2021, pp. 9391--9404.

\bibitem{gnnexplainer}
Z.~Ying, D.~Bourgeois, J.~You, M.~Zitnik, and J.~Leskovec, ``{GNNExplainer}:
  Generating explanations for graph neural networks,'' \emph{NeurIPS}, 2019.

\bibitem{xsubgraph}
H.~Yuan, H.~Yu, J.~Wang, K.~Li, and S.~Ji, ``On explainability of graph neural
  networks via subgraph explorations,'' in \emph{ICML}, 2021, pp.
  12\,241--12\,252.

\bibitem{gstarx}
S.~Zhang, Y.~Liu, N.~Shah, and Y.~Sun, ``{GStarX}: Explaining graph neural
  networks with structure-aware cooperative games,'' \emph{NeurIPS}, pp.
  19\,810--19\,823, 2022.

\bibitem{PGM}
M.~N. Vu and M.~T. Thai, ``{PGM-Explainer}: Probabilistic graphical model
  explanations for graph neural networks,'' in \emph{NeurIPS}, 2020.

\bibitem{Counter}
Z.~Huang, M.~Kosan, S.~Medya, S.~Ranu, and A.~K. Singh, ``Global counterfactual
  explainer for graph neural networks,'' in \emph{WSDM}, 2023, pp. 141--149.

\end{thebibliography}
\end{document}